# ZENN: A Thermodynamics-Inspired Computational Framework for Heterogeneous Data-Driven Modeling

**Shun Wang[a], Shun-Li Shang[b], Zi-Kui Liu[b,2], and Wenrui Hao[a,1]**

**Traditional entropy-based methods — such as cross-entropy loss in classification problems — have long been essential tools for representing the information uncertainty and physical disorder in data and for developing artificial intelligence algorithms. However, the rapid growth of data across various domains has introduced new challenges, particularly the integration of heterogeneous datasets with intrinsic disparities. To address this, we introduce a zentropy-enhanced neural network (ZENN), extending zentropy theory into the data science domain via intrinsic entropy, enabling more effective learning from heterogeneous data sources. ZENN simultaneously learns both energy and intrinsic entropy components, capturing the underlying structure of multi-source data. To support this, we redesign the neural network architecture to better reflect the intrinsic properties and variability inherent in diverse datasets. We demonstrate the effectiveness of ZENN on classification tasks and energy landscape reconstructions, showing its superior generalization capabilities and robustness—particularly in predicting high-order derivatives. In image and text classification tasks, ZENN demonstrates superior generalization by introducing a learnable temperature variable that models latent multi-source heterogeneity, allowing it to surpass state-of-the-art models on CIFAR-10/100, BBCNews, and AGNews. As a practical application in materials science, we employ ZENN to reconstruct the Helmholtz energy landscape of $Fe_3Pt$ using data generated from density functional theory (DFT) and capture key material behaviors, including negative thermal expansion and the critical point in the temperature-pressure space. Overall, this work presents a zentropy-grounded framework for data-driven machine learning, positioning ZENN as a versatile and robust approach for scientific problems involving complex, heterogeneous datasets.**

The rise of complex, heterogeneous datasets is transforming data science, enabling new capabilities across fields such as biomedical modeling, materials discovery, climate prediction, and the development of digital twins. These digital replicas of physical systems rely on the seamless integration of real-time and historical data from multiple sources, including sensors, machines, experiments, and simulations (1–3). As a powerful paradigm for empirical discovery, data-driven modeling has been widely applied to solve inverse problems in physics (4), uncover biological mechanisms (5–7), and enable precision diagnostics for complex diseases (8, 9).

A key tool in these approaches is **entropy**, a fundamental concept in thermodynamics and information theory. Entropy provides a framework for modeling uncertainty, disorder, and irreversibility across a wide range of systems, from black holes (10, 11) and ecosystems (12) to digital infrastructures and quantum materials (13–15). According to the second law of thermodynamics, internal processes in closed systems are irreversible, leading to positive entropy production (12, 16). Thermodynamic principles have increasingly influenced deep learning: for instance, neural networks have been developed to construct DFT-based potential energy surfaces (17), Boltzmann Generators embed Boltzmann entropy to predict protein structures (18), and entropy-based frameworks like Thermodynamics-Inspired Explainable Representations and EPR-Net improve interpretability and modeling of high-dimensional non-equilibrium systems (19–21). Entropy-driven objectives—such as cross-entropy loss—have been widely applied in machine learning(22–25), particularly for classification tasks (26). However, traditional entropy-based methods typically provide only macroscopic or microscopic approximations and often fail to account for internal disparities when fusing multi-source data. These limitations are particularly pronounced for real-world datasets,

## Significance Statement

The increasing availability of complex, heterogeneous datasets poses significant challenges for traditional data-driven methods, which often assume data homogeneity and fail to account for internal disparities. Quantifying entropy and its evolution in such settings remains a fundamental problem in digital twins and data science. While traditional entropy-based approaches provide useful approximations, they are limited in handling multi-source, dynamically evolving systems. To address these challenges, we introduce a **zentropy-enhanced neural network (ZENN)**—a novel framework that extends zentropy theory from quantum and statistical mechanics to data science by assigning **intrinsic entropy** to each dataset. ZENN simultaneously learns both Helmholtz energy and intrinsic entropy, enabling robust generalization, accurate high-order derivative prediction, and adaptability to heterogeneous, real-world data.

Author affiliations: [a]Department of Mathematics, Penn State University, University Park, Pennsylvania, United States of America; [b]Department of Materials Science and Engineering, Penn State University, University Park, Pennsylvania, United States of America

Author Contributions: Wenrui Hao and Zi-Kui Liu designed research; Shun Wang and Wenrui Hao contributed development of methodology; Shun Wang and Shun-Li Shang performed research; Shun Wang, Shun-Li Shang, Zi-Kui Liu and Wenrui Hao wrote the paper.

The authors have declared that no competing interests exist.

[1]To whom correspondence should be addressed. E-mail: wxh64@psu.edu.

[2]To whom correspondence should be addressed. E-mail: zxl15@psu.edu.

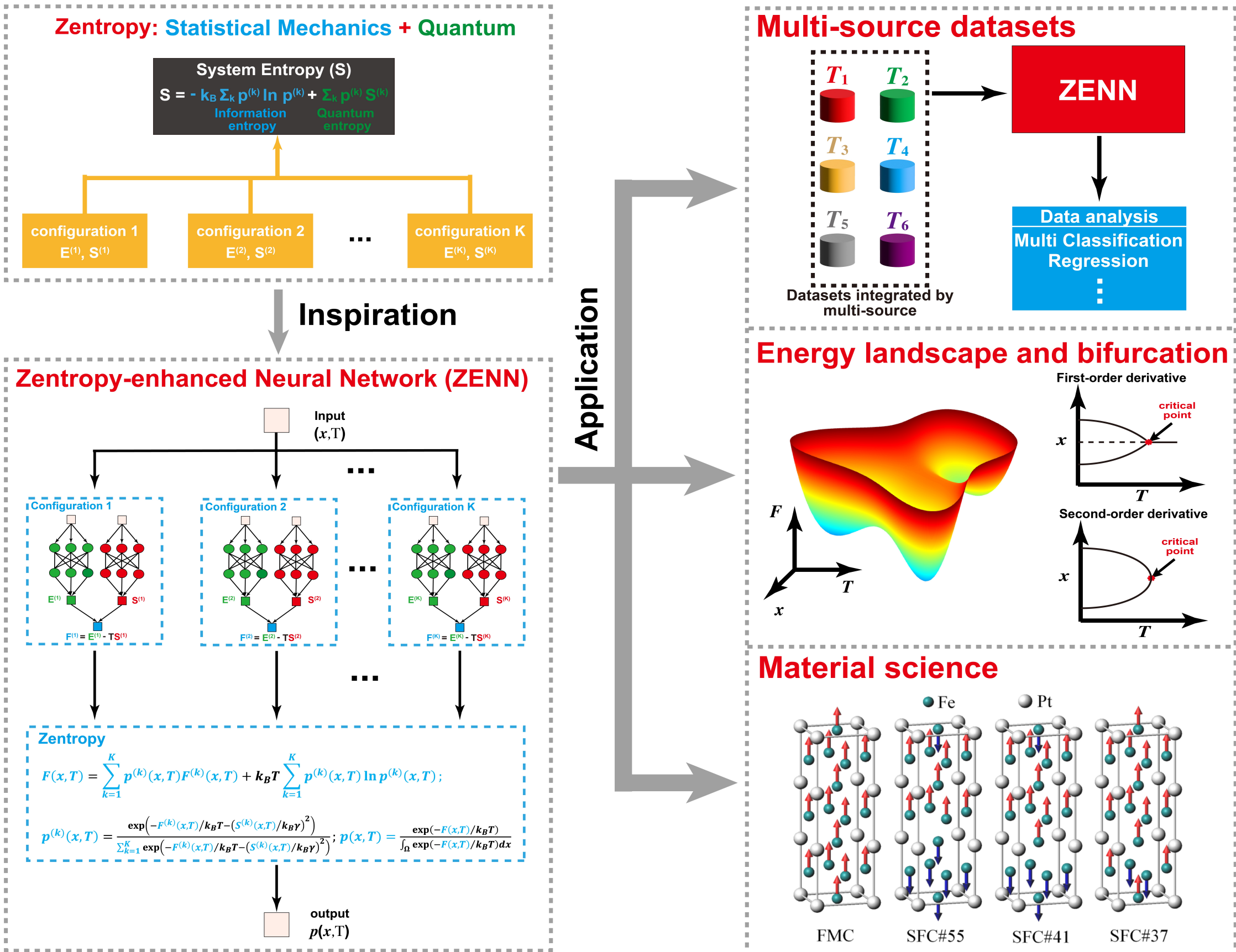

**Fig. 1. Schematic of ZENN and its applications in different areas.** Zentropy theory integrates statistical mechanics and quantum mechanics by assigning intrinsic entropy to each configuration in the system, thereby capturing internal disparities. By embedding zentropy theory into deep learning as a data-driven modeling framework, ZENN replaces the internal energy $E^{(k)}$ and entropy $S^{(k)}$ of each configuration with simple neural networks, and integrates information across all configurations through the total Helmholtz energy $F$. In this paper, ZENN has been applied to three representative tasks—multi-source data integration, energy landscape reconstruction, and inference of $Fe_3Pt$ properties—demonstrating its potential as a powerful framework that effectively bridges statistical mechanics and machine learning.

which vary in scale, resolution, measurement conditions, and statistical properties—posing challenges for conventional machine learning methods that assume data homogeneity and stationarity (27).

To overcome these challenges, **zentropy theory** introduces a thermodynamically consistent framework that incorporates **intrinsic entropy**, capturing variability across datasets due to environmental, experimental, or contextual differences (28, 29). Originally developed to predict magnetic phase diagrams in fcc-Ce (30) and $L1_2$-$Fe_3Pt$ (31), zentropy has since been applied to a wide range of materials, including strongly correlated systems (32–34), liquid and melting phases (35), plasticity (36), and superconductivity (37). Despite its success in physics, zentropy's reliance on DFT for computing configuration-level entropy has limited its application in broader data-driven domains. Nevertheless, as a foundational framework for integrating heterogeneous data, zentropy remains essential for ensuring thermodynamic consistency and enabling more accurate, comprehensive models.

Building on this foundation, we introduce a **zentropy-enhanced neural network (ZENN)** that embeds zentropy theory into deep learning to enable **entropy-aware learning from complex, heterogeneous datasets**. Unlike Shannon entropy, zentropy's thermodynamic basis captures domain-specific complexity, providing richer structure for modeling multi-source data. By incorporating zentropy into neural network design, ZENN learns both energy and intrinsic entropy components, capturing the underlying structure of heterogeneous datasets. This approach extends the applicability of zentropy beyond quantum and statistical mechanics, offering a robust framework for integrating heterogeneous, real-world data and improving predictive performance in classification, energy landscape reconstruction, and other scientific applications.

While zentropy has been widely used in forward modeling, where it plays a critical role in predicting system behavior, its integration into data-driven approaches for inverse problems remains unexplored. This paper addresses this gap by embedding zentropy theory into data science, advancing the application of entropy-based modeling within the context of data-driven techniques. More specifically, to broaden the applicability of zentropy theory beyond its traditional domain, we developed ZENN—a framework designed to integrate multi-source data, reconstruct energy landscapes, and infer system properties. We benchmarked ZENN against cross-entropy loss using state-of-the-art models for both image and text classification tasks and energy landscape reconstruction. In all cases, ZENN consistently outperformed cross-entropy loss in terms of accuracy and stability. To further validate its capabilities, ZENN was applied to the $Fe_3Pt$ compound, successfully capturing its negative thermal expansion (NTE) behavior and accurately predicting its critical point in the temperature-pressure/volume spaces. These results demonstrate ZENN's strong generalization ability and its enhanced performance in estimating high-order derivatives—an essential feature for modeling complex thermodynamic systems.

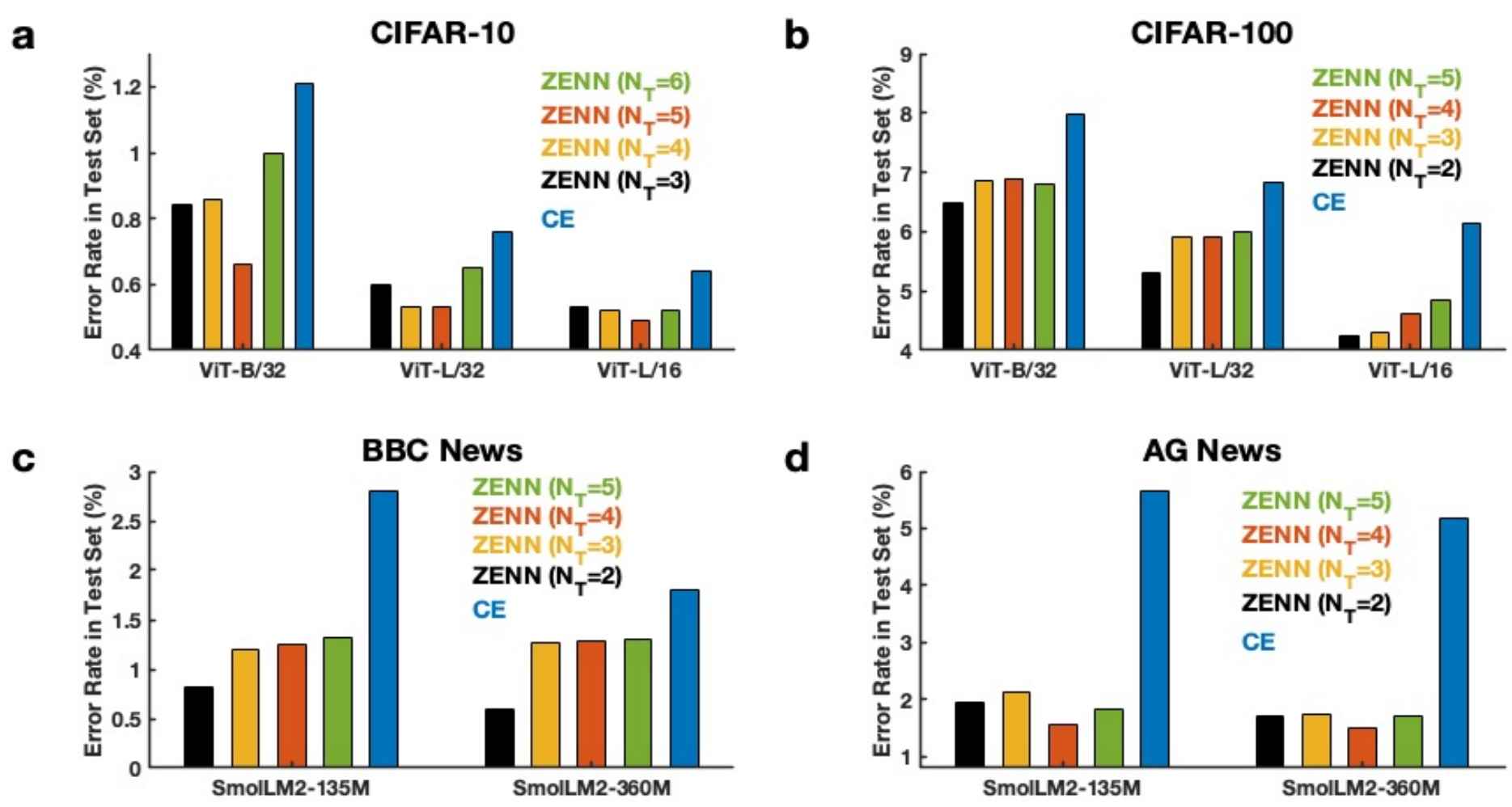

**Fig. 2. Comparison between Cross-Zentropy and Cross-Entropy both on image and text datasets.(a)** Error rate among three vision transformers as ViT-B/32, ViT-L/32, and ViT-L/16 on CIFAR-10 image dataset.**(b)** Error rate mong three vision transformers as ViT-B/32, ViT-L/32, and ViT-L/16 on CIFAR-100 image dataset.**(c)** Error rate between two language models (LMs) SmolLM2-135M, and SmolLM2-360M on BBCNews text dataset.**(d)** Error rate between SmolLM2-135M, and SmolLM2-360M on AGNews text dataset. $N_T$ represents the number of temperature modes.

## Results

**Overview of ZENN and Its Applications.** Zentropy theory states that the total entropy of a system comprises both the entropy of individual configurations and the statistical entropy across configurations. In forward problems, the Helmholtz energy $F$ of the system can be calculated if the internal energy $E^{(k)}$, entropy $S^{(k)}$ of each configuration, and the number of configurations $K$ are known (for details, see *Materials and Methods*). For materials applications, $E^{(k)}$ and $S^{(k)}$ are typically computed using DFT, while $K$ is determined by the supercell size, reflecting the system's internal degrees of freedom.

In this paper, we introduce an inverse problem by recovering $E^{(k)}$ and $S^{(k)}$ and representing them with shallow neural networks (one or two hidden layers) by fitting the data. The number of configurations, $K$—analogous to the depth of a neural network—is incrementally increased until the loss converges (Fig. 1). Once $E^{(k)}$ and $S^{(k)}$ are determined via this data-driven approach, we apply zentropy theory to compute the probability $p^{(k)}$ of each configuration and its corresponding Helmholtz energy $F^{(k)}$. This enables us to reconstruct the systematic probability density $p$ and the energy landscape $F$ using ZENN (Fig. 1). By leveraging ZENN, we analyze multi-source data, encompassing classification tasks, and reconstruct the underlying energy landscape (Fig. 1). Furthermore, we identify critical points, where systems exhibit singularity and diverge, by applying automatic differentiation to the ZENN-reconstructed energy landscape, allowing us to compute the critical temperature $T^*$ via Eq. (19).

**ZENN Outperforms Cross-entropy Both in Image and Text Classification Tasks.** We compared our proposed cross-zentropy with cross-entropy losses on both image and text datasets. For image classification, CIFAR-10 and CIFAR-100 (38) were used, with $K = 10$ and $K = 100$ configurations in ZENN, respectively, matching the number of classes in each dataset. Both the internal energy $\tilde{E}(x)$ and entropy $\tilde{S}(x)$ in ZENN are modeled using three Vision Transformer models such as ViT-B/32, ViT-L/32, and ViT-L/16 (39), where $x$ denotes the input image and $T$ is a latent variable inferred using Eq. (10)–Eq. (16). In contrast, the cross-entropy loss can be recovered by setting $\tilde{S}(x) = 0$ for image classification tasks, which corresponds to the classical softmax formulation. Additional training details and hyperparameter settings are provided in Supplementary Tables S3-S4 and Figs. S7-S12.

On both CIFAR-10 and CIFAR-100 datasets, ZENN consistently outperforms cross-entropy in test accuracy (Figs. 2a and 2b) while achieving comparable training loss (Supplementary Figs. S7 and S10). With an optimally learned temperature set, ZENN achieves lower test error rates than cross-entropy across all three Vision Transformer models, reducing the relative error by approximately 20-50% on both CIFAR-10 (Fig. 2a) and CIFAR-100 (Fig. 2b). Interestingly, when ZENN is integrated into ViT-B/32—the smallest model in the Vision Transformer family (See Supplementary Table S9), it achieves an error rate comparable to that of the large Vision Transformer model with the same token count (ViT-L/32) fine-tuned using cross-entropy.

The CIFAR series is a dataset with latent multi-source structures (40) that conventional cross-entropy loss cannot capture. ZENN overcomes this limitation through a learnable temperature variable that identifies hidden heterogeneity within the data. By inferring the posterior probability distribution over temperature assignments, ZENN robustly identifies the optimal temperature set using different models for each dataset (Figs. 2a and 2b) and reveals the likely source of each sample (See Supplementary Figs. S8 and S11). Using a one-hot encoding of the maximum a posteriori (MAP) temperature, we observe that about 60% of test samples concentrate in a dominant mode (See Supplementary Figs. S8 and S11), consistent with the strong performance of the cross-entropy baseline. When integrated with the Expectation-Maximization (EM) algorithm, ZENN uncovers multiple temperature modes (See Supplementary Figs. S8 and S11) that reflect latent heterogeneity in the dataset, thereby enhancing generalization for image classification tasks.

For the text classification task, we utilized the BBCNews (41) and AGNews (42) datasets. The number of configurations in ZENN was set to five ($K = 5$) for BBCNews, corresponding to its five classes, and four ($K = 4$) for AGNews, corresponding to its four classes. Both the internal energy $\tilde{E}(x)$ and entropy $\tilde{S}(x)$ in ZENN are modeled using SmolLM language models, including SmolLM2-135M and SmolLM2-360M (43), where $x$ denotes the input text and $T$ is a latent variable inferred using Eq. (10)–Eq. (16). As in the image classification tasks, the cross-entropy loss can be recovered by setting $\tilde{S}(x) = 0$ for text classification tasks. Additional training details and hyperparameter settings are provided in Supplementary Tables S5-S6 and Figs. S13-S18.

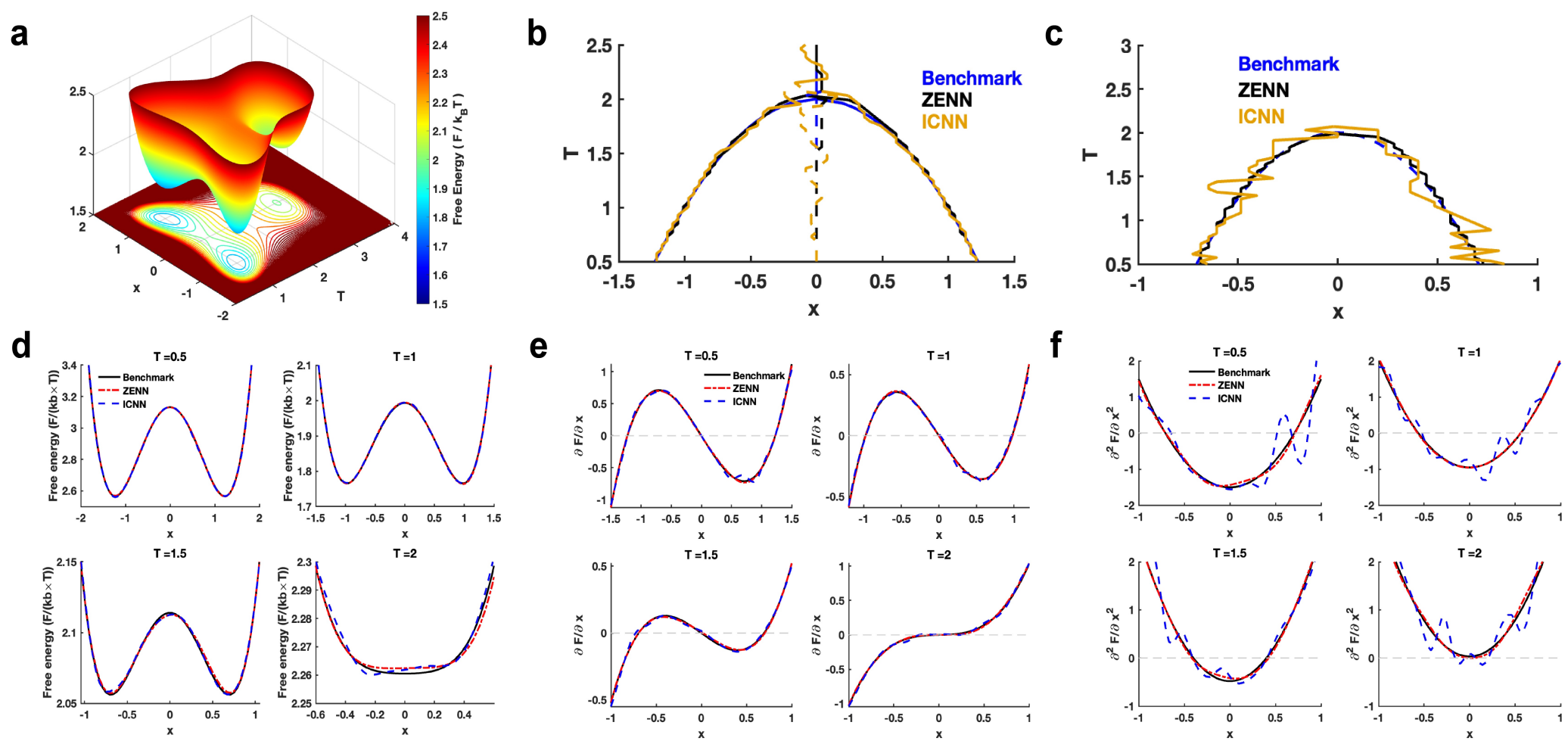

**Fig. 3. Reconstruction of the energy landscape using ZENN and ICNN. (a)** Energy landscape of the benchmark model. **(b)** Bifurcation diagram of $x$ at different values of $T$. The dashed line represents the unstable state, while the solid line indicates the metastable state. **(c)** Zero-solution line of the second-order derivative $\frac{\partial^2 F}{\partial x^2} = 0$. The blue, black, and yellow lines represent the results from the benchmark model, ZENN, and ICNN, respectively. **(d)** Helmholtz energy $F/(k_B T)$ at four different values of $T$. **(e)** First-order derivative $\frac{\partial F}{\partial x}$ at four different values of $T$. **(f)** Second-order derivative $\frac{\partial^2 F}{\partial x^2}$ at four different values of $T$. The black, blue, and red lines correspond to the benchmark model, ZENN, and ICNN, respectively. The dashed grey line indicates the zero-solution line.

On both the BBCNews and AGNews datasets, ZENN also outperforms cross-entropy in test accuracy (Figs. 2c and 2d) and achieves lower test error rates than cross-entropy across all two language models (SmolLM2), reducing the relative error by approximately 60-70% with an optimally learned temperature set. The result also showed that ZENN integrated into the smaller SmolLM2 language model (SmolLM2-135M) (See Supplementary Tables S10 and S11) achieves a lower error rate than large SmolLM2 language model (SmolLM2-360M) fine-tuned using cross-entropy (Figs. 2c and 2d).

The BBCNews, and AGNews datasets contain latent multi-source structures (42). As in the image classification tasks, ZENN infers the posterior temperature distribution to robustly determine the optimal temperature set for each dataset (Figs. 2c and 2d) and identifies the likely source of each sample (See Supplementary Figs. S14 and S17). Using a one-hot encoding of the MAP temperature, we observe that about 50% of test samples concentrate in a dominant mode (See Supplementary Figs. S14 and S17), consistent with the strong performance of the cross-entropy baseline. When integrated with the EM algorithm, ZENN also uncovers multiple temperature modes (See Supplementary Figs. S14 and S17) that reflect latent heterogeneity in the dataset, thereby improving generalization for the text classification tasks.

**ZENN Reconstructs the Energy Landscape and Robustly Identifies Bifurcation Points.** To further assess the applicability of ZENN, a benchmark model for the energy landscape was constructed. ZENN was then used to learn the configurations, reconstruct the energy landscape, and identify its bifurcation point, where the energy landscape changes. The benchmark model was defined as

$$F(x,T) = k_B T \left[ \left( \frac{x^2}{2} + \frac{T-2}{2} \right)^2 + \frac{\left( (T-2)^2 - 1 \right)^2}{2} \right],$$

where $k_B$ is the Boltzmann constant. As shown in Fig. 3a, when $T = 1$, the model has double wells. As $T$ increases to 3, the double well degenerates into a single well. At $T = 2$, a saddle curve emerges, indicating that $T = 2$ is a bifurcation point for this benchmark model. This observation can be validated using both the first-order condition,

$$\left. \frac{\partial F}{\partial x} \right|_{T=2} = 0,$$

and the second-order condition,

$$\left. \frac{\partial^2 F}{\partial x^2} \right|_{T=2} = 0.$$

In ZENN, the internal energy $E^{(k)}(x,T)$ and entropy $S^{(k)}(x,T)$ for each configuration are modeled by neural networks with two hidden layers of eight neurons each, taking $(x,T)$ as input. The number of configurations in ZENN was set to six ($K = 6$). The Input convex neural networks (ICNNs) (See details in Supplementary Section 2) (44) was configured with four hidden layers, each containing 48 neurons, ensuring the total number of neurons matched that of ZENN. Parameters for both ZENN and ICNN were estimated using the Jensen–Shannon divergence (Eq. (17)). First- and second-order derivatives were computed via automatic differentiation in both ZENN and ICNN (Figs. 3b–c).

The result of the Helmholtz energy $F/(k_B T)$ at $T = 2$ shows a significant difference between ZENN and ICNN (Fig. 3d). The bifurcation diagram of $x$ at different values of $T$, reconstructed by ZENN and ICNN (Fig. 3b), shows that the first-order derivative from ZENN aligns more closely with the benchmark model than that from ICNN. At all four different values of $T$, the first-order derivative of ICNN exhibits fluctuations, though these have minimal impact on the bifurcation diagram (Fig. 3e). However, these fluctuations lead to inaccurate predictions of the second-order derivative. The zero-contour line of the second-order derivative (Fig. 3c) demonstrates that ZENN provides more robust high-order derivative predictions than ICNN. At all four different values of $T$, the second-order derivative predicted by ZENN aligns well with the benchmark model, whereas ICNN fails to predict both the second-order derivative and the correct number of zero crossings (Fig. 3f). These results underscore ZENN's robustness in high-order derivative estimation and its potential for identifying critical points where the energy landscape changes. While ICNNs with L2 fitting are well-known to perform poorly in approximating derivatives, Sobolev training has been proposed to address this challenge by leveraging derivative information from the data (45). However, ZENN can effectively approximate derivatives using only the L2 fitting setup, demonstrating its superior ability to capture high-order derivatives without the need for additional derivative data.

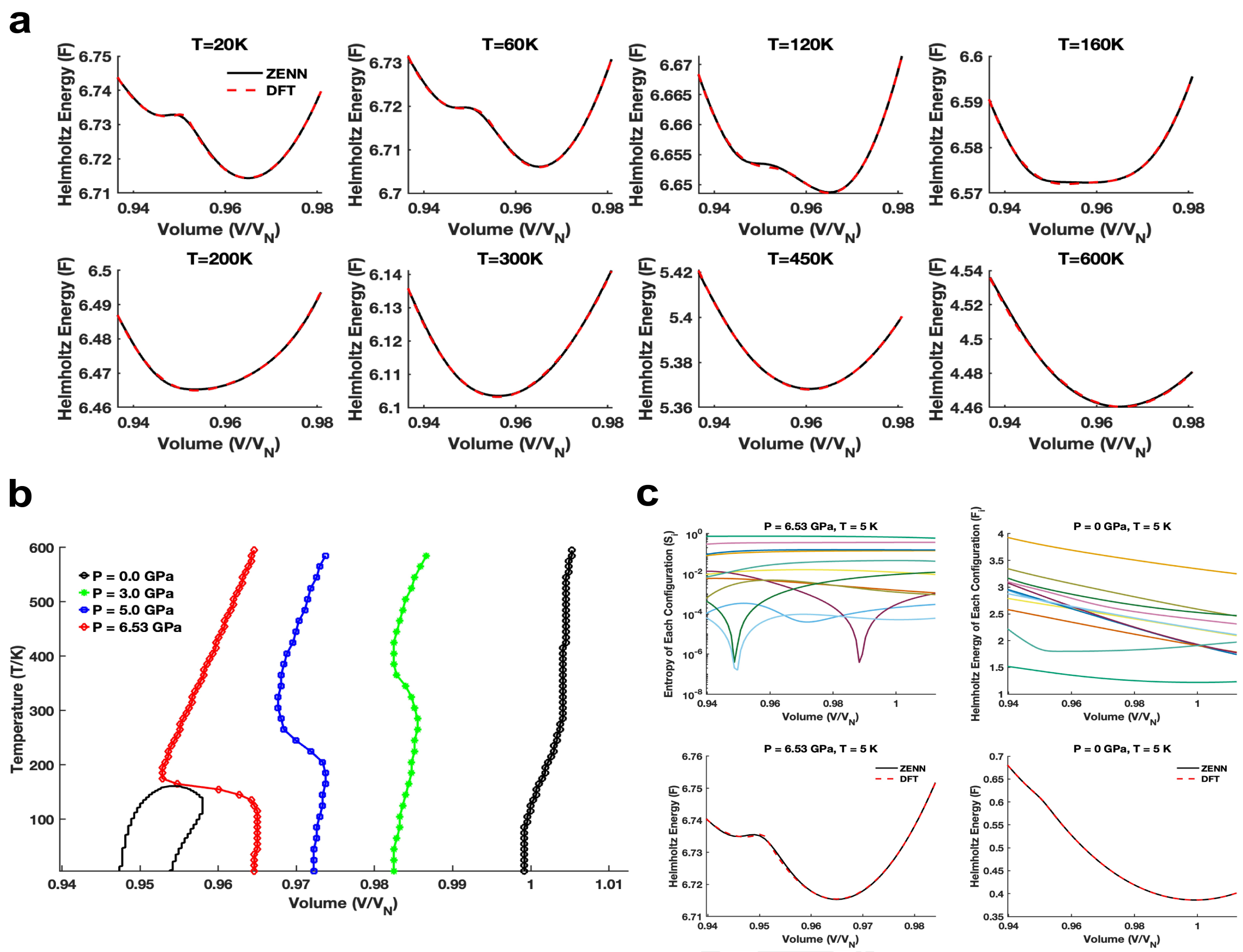

**Fig. 4. ZENN-Based Identification of NTE and the Critical Temperature of $Fe_3Pt$.** **(a)** Helmholtz energy of $Fe_3Pt$ as a function of volume $V$ at different temperatures $T$, under a pressure of 6.53 GPa. **(b)** Predicted $T$-$V$ phase diagram of $Fe_3Pt$, showing isobaric volume curves, where the volume $V$ is normalized to $V_N$ at 298 K and 1 atm. **(c)** Entropy and Helmholtz energy of each configuration at $T = 5\,\mathrm{K}$. Colored lines represent the twelve configurations. Solid and dashed lines indicate the Helmholtz energy of $Fe_3Pt$ predicted by ZENN and DFT, respectively.

Additionally, we evaluated the approximation capacity of ZENN on a more complex energy landscape to further demonstrate its robustness. Specifically, we constructed a two-dimensional energy landscape featuring three wells and illustrated ZENN's effectiveness in accurately capturing this landscape in Supplementary Figs. S3 and S4.

**ZENN Accurately Captures Negative Thermal Expansion and the Critical Point of $Fe_3Pt$.** We further applied ZENN to the Helmholtz energy data of $Fe_3Pt$ generated by DFT to assess its ability to predict material properties. A 12-atom $1 \times 1 \times 3$ supercell was employed, where nine Fe atoms possess spin-up and/or spin-down orientations, leading to $2^9 = 512$ total configurations, of which 37 are symmetry-independent. The DFT-based quasi-harmonic approximation (QHA), implemented in the VASP code (46), was used to compute the independent $F^{(k)}$ values, with full details provided in previous work (47). From these $F^{(k)}$ values, the total Helmholtz energy $F$ of $Fe_3Pt$ was calculated (Figs. 4a and S1 in Supplementary Materials; see Supplementary Materials for details).

In ZENN, both the internal energy $E^{(k)}(x, T)$ and entropy $S^{(k)}(x, T)$ of each configuration were modeled using neural networks with two hidden layers, each containing eight neurons. The variable $x$ denotes the volume of $Fe_3Pt$ and is one-dimensional. The number of configurations in ZENN was set to twelve ($K = 12$). By incorporating a convexity constraint into the Jensen–Shannon divergence loss function (Eq. (20)), ZENN's predicted Helmholtz energy shows remarkable agreement with the DFT data across all temperatures (Fig. 4a and Supplementary Fig. S1).

Using ZENN, we computed the isobaric volume curves $\frac{\partial F}{\partial V} = -p$, the zero-solution line of the second-order derivative $\frac{\partial^2 F}{\partial V^2} = 0$, and the critical temperature $T^*$ via Eq. (19). Under atmospheric pressure, ZENN accurately captures the negative thermal expansion (NTE), which becomes more pronounced with increasing pressure (Fig. 4b). At 6.53 GPa, ZENN predicts a critical temperature of 161 K (Fig. 4b), which aligns closely with the DFT-computed range of 160–165 K (Supplementary Fig. S2). These results are consistent with previous experimental and DFT-based studies (31, 48), demonstrating ZENN's strong potential for predicting material properties.

It is important to note that some $S^{(k)}$ curves at 5 K and 6.53 GPa from ZENN (Fig. 4c) decrease with increasing volume and even exhibit sharp drops. These irregularities, when compared with DFT-based results, arise from certain combined configurations employed in ZENN. The intersection of Helmholtz energies for these combined configurations at specific volumes results in sharp fluctuations of $p^{(k)}$ values as well as sharp drops of $S^{(k)}$ values, as observed by ZENN at low temperatures (e.g., below 100 K, see Supplementary Fig. S6). For example, the sudden drop of $S^{(k)}$ around the normalized volume of 0.99 is primarily due to the abrupt predominance of ferromagnetic configuration (having the lowest entropy) in the combined configuration in ZENN. However, the total entropy from both ZENN and DFT shows similar trends as a function of volume (see Fig. 4b and Fig. 5 in (49)), exhibiting a maximum point corresponding to NTE in $Fe_3Pt$ (49). Additionally, one of the individual $S^{(k)}$ curves shows a large slope change with respect to volume (near 2 eV/atom in Fig. 4c) also due to the combination of different configurations utilized in ZENN. The Helmholtz energy of all configurations remains convex (Fig. 4c) at 5 K and 0 GPa. At both 0 GPa and 6.53 GPa, the total Helmholtz energy at 5 K predicted by ZENN shows excellent agreement with DFT results (Fig. 4c), further reinforcing ZENN's applicability in materials science.

## Discussions

In this study, we introduce and demonstrate ZENN, a data-driven framework grounded in zentropy theory that integrates neural networks with statistical mechanics to approximate energy landscapes, predict class probabilities, and identify critical transitions in material systems. By parameterizing the energy and entropy of individual configurations through compact neural networks, ZENN provides a systematic and interpretable approach for reconstructing Helmholtz energy landscapes and capturing the complex thermodynamic behaviors of systems with multiple configurations. In contrast to conventional DNNs, where each layer is fully connected, ZENN models each configuration with a small, independent neural network, with the number of configurations roughly corresponding to the number of DNN layers. This design replaces strong nonlinearity across layers with physically grounded connections via thermodynamic principles, enhancing both robustness and derivative accuracy.

The entropy fluctuation constraint $\sum_{k=1}^{n} p^{(k)}(S^{(k)})^2 = C_2$ is introduced to quantify the variation of entropy across individual configurations. This constraint is also essential for distinguishing entropy from energy within the model. From a statistical perspective, this term corresponds to a second-order moment (i.e., the variance-like measure of entropy), whereas $\sum_{k=1}^{n} p^{(k)}E^{(k)} = C_1$ represents the first-order moment, i.e., the expected energy. According to the derivation of $p^{(k)}$, the associated Lagrange multiplier $\gamma$ for the entropy fluctuation constraint is also expected to possess a physical interpretation — analogous to how temperature emerges as the Lagrange multiplier associated with the energy constraint.

Our results across image and text classification tasks, energy landscape reconstruction, and property prediction for $Fe_3Pt$ — demonstrate that ZENN consistently outperforms traditional cross-entropy loss. Specifically, both in image and text classification tasks , ZENN demonstrates superior generalization by incorporating intrinsic entropy contributions—a feature absent in conventional cross-entropy (Fig. 2). The CIFAR datasets are inherently multi-source, constructed from tiny images gathered through diverse search engines and keywords, leading to variations in image style and compression within each category. For example, results from CIFAR-10-Warehouse (40) have noted the presence of latent multi-source structures that traditional cross-entropy loss fails to capture. The BBCNews and AGNews datasets contain inherent multi-source signals, as noted in their dataset documentation (42). By introducing a learnable temperature variable, ZENN effectively distinguishes these hidden data sources and integrates their heterogeneity (See Supplementary Figs. S8, S11, S14 and S17), resulting in improved accuracy. Moreover, within the same model family, a small model with ZENN could outperform a larger model fine-tuned with cross-entropy (Fig. 2), demonstrating that ZENN handles data heterogeneity more effectively than simply scaling up parameters.

More broadly, the cross-zentropy loss employed by ZENN simultaneously accounts for both energy and entropy, reducing overfitting and improving prediction reliability, particularly in heterogeneous or imbalanced datasets. This aligns with recent efforts in physics-informed machine learning, such as Sobolev training for derivative estimation (45) and Boltzmann generators for energy-based sampling, yet ZENN uniquely bridges these advances by explicitly modeling entropy fluctuations, enabling robust multi-source data integration.

In reconstructing energy landscapes, ZENN robustly predicts high-order derivatives and identifies bifurcation points that are typically inaccessible to conventional ICNNs (Figs. 3c, f). While ICNNs often require explicit derivative supervision to capture such behaviors, ZENN's thermodynamic foundation ensures stable and physically consistent predictions of quantities like $\frac{\partial^2 F}{\partial x^2}$. This capability is particularly valuable in materials science, where subtle curvature changes in free energy landscapes govern phase transitions, such as magnetic ordering or negative thermal expansion. Notably, ZENN accurately predicted the critical temperature for $Fe_3Pt$, closely matching DFT and experimental results (Fig. 4b).

Beyond materials modeling, the ZENN framework is broadly applicable to other fields where energy landscapes or heterogeneous data distributions are central. Potential applications include protein folding in biophysics, climate data integration from satellite and ground measurements, and digital twins that fuse multi-resolution, real-time sensor data. ZENN's ability to quantify and leverage intrinsic entropy differences offers a unique advantage in these contexts, ensuring consistent predictions across disparate datasets.

A key distinction between zentropy theory and classical Gibbs entropy lies in their treatment of configurational entropy. While Gibbs entropy adopts a top-down perspective based on probability distributions, zentropy theory integrates quantum and statistical mechanics to compute the total entropy from configuration-specific contributions, thereby im-proving the accuracy of phase-transition predictions (28, 29). In forward modeling, zentropy theory requires evaluating all configurations—for example, 512 in $Fe_3Pt$—to capture behaviors like negative thermal expansion. In contrast, ZENN, as an inverse modeling framework, reconstructs the same energy landscape and predictive fidelity using as few as 12 configurations, demonstrating remarkable data efficiency.

Despite its advantages, ZENN's current implementation requires predefining the number of configurations $K$, potentially limiting scalability for ultra-high-dimensional systems. Future extensions could explore adaptive or hierarchical configuration structures, or coupling ZENN with generative models and active learning frameworks to efficiently explore large configuration spaces. Additionally, while a convexity constraint was employed here for physical consistency in $Fe_3Pt$, other systems (e.g., glasses or non-equilibrium materials) may require alternative regularization schemes. Extending ZENN to dynamic systems with time-dependent entropy production would also unlock new applications in non-equilibrium thermodynamics and real-time monitoring.

In summary, ZENN represents a significant advancement in data-driven modeling by embedding thermodynamic laws directly into deep learning architectures. Its demonstrated success in classification, energy landscape reconstruction, and materials property prediction highlights the transformative potential of integrating domain-specific physics with modern machine learning. As complex, heterogeneous datasets become increasingly ubiquitous, ZENN's interpretability, data efficiency, and robust handling of high-order derivatives position it as a powerful tool for accelerating discovery and design in materials science and beyond.

## Materials and Methods

**Zentropy through integration of Statistical and Quantum Mechanics.** *Zentropy theory* (49) is fundamentally a multiscale entropy framework (29, 49, 50), where the "z" denotes the partition function, derived from the German word *zustandssumme*, meaning "sum over states." This highlights that zentropy theory involves summing the entropies across multiple hierarchical levels. By integrating quantum mechanics and statistical mechanics (Fig. 1), zentropy theory (29) constructs a coarse-grained representation of entropy as follows:

$$S = \sum_{k=1}^{K} p^{(k)} S^{(k)} - k_B \sum_{k=1}^{K} p^{(k)} \ln p^{(k)} \quad [1]$$

where $S^{(k)}$ ($S^{(k)} \geq 0$) is the entropy of the $k$-th configuration. This is related to the degrees of freedom in vibrational frequencies of atoms in quantum mechanics (49). Eq. (1) can also be derived from the recursive property of entropy as shown recently (51). Here $k_B$ is the Boltzmann constant and $p^{(k)}$ is the probability of the $k$-th configuration, which is inferred by the maximum entropy principle under the following three constraints:

$$\begin{cases} \sum_{k=1}^{n} p^{(k)} = 1 & \text{(Normalization)} \\ \sum_{k=1}^{n} p^{(k)} E^{(k)} = C_1 & \text{(Total internal energy constraint)} \\ \sum_{k=1}^{n} p^{(k)} \left(S^{(k)}\right)^2 = C_2 & \text{(Entropy fluctuation constraint)} \end{cases}$$

where $C_1$ and $C_2$ are both constants. Whereas the first two constraints are standard in entropy-based modeling, we introduce the third constraint to quantify the fluctuation of entropy across individual configurations, which effectively reflects the supercell size in DFT calculations. Thus, the optimization problem is summarized as:

$$\max_{p^{(k)}} \; S = \sum_{k=1}^{K} p^{(k)} S^{(k)} - k_B \sum_{k=1}^{K} p^{(k)} \ln p^{(k)} \quad [2]$$

s.t.

$$\sum_{k=1}^{K} p^{(k)} = 1, \quad \sum_{k=1}^{K} p^{(k)} E^{(k)} = C_1, \quad \sum_{k=1}^{K} p^{(k)} \left(S^{(k)}\right)^2 = C_2$$

By introducing Lagrange multipliers $\lambda_i$ ($i = 1, 2, 3$), the corresponding Lagrangian is given by:

$$\begin{aligned} \mathcal{L}(p^{(k)}) = &\sum_{k=1}^{K} p^{(k)} S^{(k)} - k_B \sum_{k=1}^{K} p^{(k)} \ln p^{(k)} \\ &- \lambda_1 \left( \sum_{k=1}^{K} p^{(k)} - 1 \right) - \lambda_2 \left( \sum_{k=1}^{K} p^{(k)} E^{(k)} - C_1 \right) \\ &- \lambda_3 \left( \sum_{k=1}^{K} p^{(k)} \left(S^{(k)}\right)^2 - C_2 \right) \end{aligned} \quad [3]$$

Thus, the optimal solution satisfies:

$$\frac{\partial \mathcal{L}}{\partial p^{(k)}} = S^{(k)} - k_B \left( \ln p^{(k)} + 1 \right) - \lambda_1 - \lambda_2 E^{(k)} - \lambda_3 \left(S^{(k)}\right)^2 = 0 \quad [4]$$

which implies

$$p^{(k)} = \underbrace{\exp\left( -\frac{\lambda_1 + k_B}{k_B} \right)}_{\text{Normalization}} \underbrace{\exp\left( \frac{S^{(k)} - \lambda_2 E^{(k)}}{k_B} \right)}_{\text{Total internal energy}} \underbrace{\exp\left( -\frac{\lambda_3 \left(S^{(k)}\right)^2}{k_B} \right)}_{\text{Fluctuation of intrinsic entropy}} \quad [5]$$

Due to the nondimensionalization of $p^{(k)}$, we set $\lambda_2 = \frac{1}{T}$ and $\lambda_3 = \frac{1}{\gamma^2 k_B}$, where $\gamma$ is a tunable hyperparameter. Since the Helmholtz energy of the $k$-th configuration is given by $F^{(k)} = E^{(k)} - TS^{(k)}$, the probability $p^{(k)}$ can be rewritten as:

$$\begin{aligned} p^{(k)} &= \frac{1}{Z} \exp\left( -\frac{F^{(k)}}{k_B T} - \left( \frac{S^{(k)}}{\gamma k_B} \right)^2 \right), \\ Z &= \sum_{k=1}^{K} \exp\left( -\frac{F^{(k)}}{k_B T} - \left( \frac{S^{(k)}}{\gamma k_B} \right)^2 \right). \end{aligned} \quad [6]$$

Thus, the total Helmholtz energy of system F is expressed as

$$\begin{aligned} F &= E - TS \\ &= \sum_{k=1}^{K} p^{(k)} E^{(k)} - T \left( \sum_{k=1}^{K} p^{(k)} S^{(k)} - k_B \sum_{k=1}^{K} p^{(k)} \ln p^{(k)} \right) \\ &= \sum_{k=1}^{K} p^{(k)} F^{(k)} + k_B T \sum_{k=1}^{K} p^{(k)} \ln p^{(k)} \end{aligned} \quad [7]$$

From Eq. (6), the partition function of the $k$-th configuration can be defined as $Z^{(k)} = \exp\left(-\frac{F^{(k)}}{k_BT} - \left(\frac{S^{(k)}}{\gamma k_B}\right)^2\right)$, which reduces to the standard definition in the forward-modeling of the zentropy theory when $\gamma k_B \gg S^{(k)}$, corresponding to relatively small fluctuations of the intrinsic entropy among configurations, as demonstrated for $YNiO_3$ (36). This formulation aligns with the classic principles of statistical mechanics developed by Gibbs, who considered systems as "of the same nature, but differing in the configurations and velocities which they have at a given instant" (52). In the present work for $Fe_3Pt$, a value $\gamma = 5$ was used.

**Cross-zentropy Loss Function in Classification Tasks.** The cross-entropy loss function is widely used in classification problems (26). We extend this framework to the zentropy setup and introduce the cross-zentropy loss function. Specifically, let $\mathbf{y}_j \in \mathbb{R}^K$ denote the true one-hot encoded label, $\mathbf{p}_j \in \mathbb{R}^K$ represent the predicted probability for the $j^{th}$ sample, and $K$ is the total number of classes. The loss function of cross-entropy, $l_{\mathrm{CE}}$, is defined as:

$$l_{\mathrm{CE}} = \frac{1}{M}\sum_{j=1}^{M}\langle \mathbf{y}_j, -\ln \mathbf{p}_j\rangle \qquad [8]$$

where $\langle\cdot\rangle$ denotes the inner product, and $M$ is the number of samples. This loss function quantifies the difference between the true labels and the predicted probabilities, penalizing incorrect classifications by increasing the loss value when the predicted probability diverges significantly from the true label.

To transition from cross-entropy to cross-zentropy loss, we introduce a modification to account for the intrinsic entropy approximation. Using Eq. (6), we modify the cross-entropy framework with zentropy theory, where the Helmholtz energy of the $k$-th configuration is given by $F^{(k)} = E^{(k)} - TS^{(k)}$. Then the cross-zentropy loss function, $l_{\mathrm{CZ}}$, is defined as:

$$l_{\mathrm{CZ}}(\tilde{E},\tilde{S}) = \frac{1}{M}\sum_{j=1}^{M}\left\langle \mathbf{y}_j,\ \frac{\tilde{E}-T_j\tilde{S}}{k_BT_j} + \left(\frac{\tilde{S}}{\gamma k_B}\right)^2 + \ln Z\right\rangle \qquad [9]$$

$$Z = \sum_{k=1}^{K}\exp\left(-\frac{F^{(k)}}{k_BT_j} - \left(\frac{S^{(k)}}{\gamma k_B}\right)^2\right)$$

where $\tilde{E} = [E^{(1)}, E^{(2)}, \ldots, E^{(K)}]$ denotes the internal energy of each class, $\tilde{S} = [S^{(1)}, S^{(2)}, \ldots, S^{(K)}]$ denotes the entropy of each class, $Z$ is chosen such that $\sum_{k=1}^{K} p^{(k)} = 1$, and $K$ is the number of classes.

To identify multi-source structures within the dataset via the temperature variable, we incorporate the Expectation-Maximization (EM) algorithm (53) into the Cross-Zentropy loss, as detailed below:

$$L(\theta_E,\theta_S;x,T) = \frac{1}{M}\sum_{j=1}^{M}\sum_{i=1}^{N_T} q_j^{(\omega)}(T_i)\cdot \mathrm{CZ}(j,i), \qquad [10]$$

where $q_j^{(\omega)}(T_i)$ denotes the posterior probability of temperature $T_i$ for the $j$-th sample, and $\mathrm{CZ}(j,i)$ is the cross-zentropy at $T_i$. Specifically:

$$q_j^{(\omega)}(T_i) = \frac{\exp\left(-\omega\cdot \mathrm{CZ}(j,i)\right)}{\sum_{i=1}^{N_T}\exp\left(-\omega\cdot \mathrm{CZ}(j,i)\right)}, \qquad [11]$$

$$\mathrm{CZ}(j,i) = -\log\left(p_\theta(x_j,y_j,T_i)\right) = -\log\left(\sum_{k=1}^{K} y_{jk}\cdot p^{(k)}(x_j,T_i)\right), \qquad [12]$$

where $y_{jk}$ is the $k$-th component of the one-hot label $y_j$. The configuration-wise probability $p^{(k)}(x_j,T_i)$ is defined as:

$$p^{(k)}(x_j,T_i) = \frac{1}{Z(x_j)}\exp\left(-\frac{E^{(k)}(x_j;\theta_E) - T_iS^{(k)}(x_j;\theta_S)}{k_BT_i} - \left(\frac{S^{(k)}(x_j;\theta_S)}{\gamma k_B}\right)^2\right), \qquad [13]$$

with partition function:

$$Z(x_j) = \sum_{k=1}^{K}\exp\left(-\frac{E^{(k)}(x_j;\theta_E) - T_iS^{(k)}(x_j;\theta_S)}{k_BT_i} - \left(\frac{S^{(k)}(x_j;\theta_S)}{\gamma k_B}\right)^2\right). \qquad [14]$$

Here, $\omega$ is a temperature selection parameter: $\omega = 5.0$ is used in training (emphasizing robustness via worst-case posterior), and $\omega = 1$ is used in testing (reflecting true posterior belief over $T$). Given a fixed temperature set size $N_T$, we employ alternating optimization to jointly update network parameters and temperature values:

$$\{\theta_E^{(t+1)},\theta_S^{(t+1)}\} = \arg\min_{\theta_E,\theta_S} L(\theta_E,\theta_S;x,T^{(t)}), \qquad [15]$$

$$T^{(t+1)} = \arg\min_{T} L(\theta_E^{(t+1)},\theta_S^{(t+1)};x,T). \qquad [16]$$

**Algorithm 1** ZENN combined with EM Algorithm

- **Initialize:** parameters $\theta_E^{(0)}$, $\theta_S^{(0)}$, temperature set $T^{(0)}$
- **for** $t = 0$ to max_epochs **do**
  - **E-step:** For each sample $x_j$, compute posterior over temperature $T_i$:
  $$q_j^{(\omega)}(T_i) = \frac{\exp(-\omega\,\mathrm{CZ}(j,i))}{\sum_{i=1}^{N_T}\exp(-\omega\,\mathrm{CZ}(j,i))}$$
  - **M-step (model parameters):** Update $\theta_E^{(t+1)}$, $\theta_S^{(t+1)}$ by minimizing:
  $$L(\theta_E,\theta_S;x,T^{(t)})$$
  - **M-step (temperature):** Update $T^{(t+1)}$ by minimizing:
  $$L(\theta_E^{(t+1)},\theta_S^{(t+1)};x,T)$$

**Jensen-Shannon divergence and Identification of critical point.** When the data or distributions are continuous and defined over continuous variables, we employ Jensen-Shannon divergence in conjunction with ZENN to extend the framework. Specifically, the loss function is defined using the

Jensen-Shannon divergence as follows:

$$\min_{\{\theta_E^{(k)},\,\theta_S^{(k)}\}_{k=1}^K} \mathcal{L}\left(\{\theta_E^{(k)},\,\theta_S^{(k)}\}_{k=1}^K;\, x, T\right) = \frac{1}{2}\,\mathrm{KL}(P\|M) + \frac{1}{2}\,\mathrm{KL}(Q\|M), \quad \text{where } M = \frac{P+Q}{2} \qquad [17]$$

where $P$ is the probability density of experimental data and $Q$ is the probability density computed by the zentropy theory with $S^{(k)}(x,T)$ and $E^{(k)}(x,T)$. The parameters $\theta_S^{(k)}$ and $\theta_E^{(k)}$ are associated with the neural networks modeling $S^{(k)}(x,T)$ and $E^{(k)}(x,T)$, respectively. The Kullback–Leibler divergence (KL) is defined as follows:

$$\begin{aligned} \mathrm{KL}(P\|M) &= \int_\Omega P(x,T)\ln P(x,T)\,dx \\ &\quad - \int_\Omega P(x,T)\ln M(x,T)\,dx, \\ \mathrm{KL}(Q\|M) &= \int_\Omega Q(x,T)\ln Q(x,T)\,dx \\ &\quad - \int_\Omega Q(x,T)\ln M(x,T)\,dx. \end{aligned} \qquad [18]$$

**Identification of critical points.** To identify the critical point of the energy landscape $F(x,T)$ using zentropy theory, we define the following zero-eigenvalue problem:

$$B(x^*, T^*, \xi) = \begin{cases} \nabla F(x^*, T^*) = 0, \\ \nabla^2 F(x^*, T^*)\,\xi = 0, \\ \|\xi\| = 1, \end{cases} \qquad [19]$$

where $\xi$ is the normalized eigenvector corresponding to the zero eigenvalue of the Hessian matrix $\nabla^2 F(x,T)$ of the energy landscape. This system allows us to identify the critical temperature $T^*$ at which the energy landscape undergoes a stability change (54, 55).

**Convexity constraint to Helmholtz energy of each configuration.** Since each configuration of $Fe_3Pt$ has a unique minimum on its Helmholtz energy (56), we introduce a convexity constraint on the Helmholtz energy for each configuration. This modification alters the loss function (Eq. (20)) as follows:

$$\min_{\{\theta_E^{(k)},\theta_S^{(k)}\}_{k=1}^K} \tilde{\mathcal{L}}\left(\{\theta_E^{(k)},\theta_S^{(k)}\}_{k=1}^K;\, V, T\right) = \mathcal{L}\left(\{\theta_E^{(k)},\theta_S^{(k)}\}_{k=1}^K;\, V, T\right) + \lambda \sum_{k=1}^K \int_\Omega \mathrm{ReLU}\left(-\frac{\partial^2 F^{(k)}}{\partial V^2}\right) dV \qquad [20]$$

$$\mathrm{ReLU}(x) = \begin{cases} x, & x \geq 0, \\ 0, & x < 0 \end{cases}$$

where $\lambda$ is the regularization term and $V$ denotes volume. With a fixed value of $\lambda$, we performed 30,000 epochs and minimized the loss. The optimal value of $\lambda$ that minimizes the loss as much as possible was found to be $10^{-4}$ (57) (Supplementary Fig. S5).

According to the zentropy framework, the frame-indifference constraint can be satisfied at the level of each configuration $k$, i.e., $E_k(Qx+b) = E_k(x)$ and $S_k(Qx+b) = S_k(x)$, for any orthogonal matrix $Q \in \mathbb{R}^{n\times n}$ with $\det(Q) = 1$, and any translation vector $b \in \mathbb{R}^n$. To enforce this symmetry in our model, EGNN-based architectures (58) can be incorporated into ZENN to ensure frame-indifference by design.

**Data and code availability.** All relevant data are within the manuscript and its Supporting Information files. The public datasets are used in this study. Source codes and data have been deposited on the GitHub repository (https://github.com/WilliamMoriaty/ZENN). All simulations were performed on a server equipped with eight NVIDIA A100 GPUs.

**Acknowledgments.** This research supported by National Institute of General Medical Sciences through grant 1R35GM146894 (SW and WH), U.S. Department of Energy (DOE) through Grant No. DE-SC0023185, and the Endowed Dorothy Pate Enright Professorship at College of Earth and Mineral Science at the Pennsylvania State University (SLS and ZKL).

# ZENN: A Thermodynamics-Inspired Computational Framework for Heterogeneous Data-Driven Modeling

Shun Wang[1], Shun-Li Shang[2], Zi-Kui, Liu[2,*] Wenrui Hao[1,*]

## Content

## Section 1 Details of first-principles calculations and first-principles thermodynamics

As detailed in (1), all DFT-based first-principles calculations of $Fe_3Pt$ were performed by the Vienna *Ab initio* Simulation Package (VASP) (2). The ion-electron interaction was described by the projector augmented wave (PAW) method (3) and the exchange-correlation functional was described by the generalized gradient approximation (GGA) developed by Perdew, Burke, and Ernzerhof (PBE) (4). Regarding electron configurations, fourteen valence electrons ($3p^6 3d^7 4s^1$) were employed for Fe and ten ($5d^9 6s^1$) for Pt. During VASP calculations, a plane wave cutoff energy of 381 eV was used for structural relaxations by means of the Methfessel-Paxton method (5), and final calculations of total energy and electronic density of states were performed by the tetrahedron method with a Blöchl correction (6) using a plane wave cutoff energy of 520 eV. The employed *k*-point meshes for the 12-atom $Fe_3Pt$ configurations were ($14\times14\times4$) and the self-consistency of total energy was converged to at least $10^{-5}$ eV/atom. Due to magnetic nature of Fe, all DFT-based calculations were performed by the spin polarization calculations.

Helmholtz energy for configuration $k$, $F^k$, can be predicted by density functional theory (DFT) based quasiharmonic approximation (QHA) with the contributions from vibrations $F^k_{vib}(V,T)$, thermal electrons $F^k_{el}(V,T)$, and the static total energy at 0 K, $E^k(V)$ (1, 7),

$$F^k(V,T) = E^k(V) + F^k_{vib}(V,T) + F^k_{el}(V,T) \ (S1)$$

$E^k(V)$ of configuration *k* at 0 K can be determined by fitting the DFT calculated energy-volume (*E*-*V*) data points using a four-parameter Birch-Murnaghan equation of state (EOS) (7),

$$E^k(V) = a_1 + a_2V^{-\frac{2}{3}} + a_3V^{-\frac{4}{3}} + a_4V^{-2} \ (S2)$$

where $a_1$, $a_2$, $a_3$, and $a_4$ are fitting parameters. Equilibrium properties from this EOS fitting include the equilibrium energy $E_0$, volume $V_0$, bulk modulus $B_0$, and the pressure derivative of bulk modulus $B'$. Usually, eight reliable data points were employed for each EOS fitting in the present work, and the fitted equilibrium properties for these 37 independent configurations are shown in Table S1. To gain Helmholtz energy of each configuration *k*, we used the DFT-based QHA with vibrational contributions from the Debye model and thermal electron contributions from the Mermin statistics (7).

## Section 2 The architecture of Input Convex Neural Network

Let the input be $x \in R^d$, the number of layers be $L$, and the hidden layer dimension of each layer be $h$. We initialize the state as $z_0 = 0$, and the update rule for the $i$-th hidden layer is given by:

$$z_{i+1} = \phi\left(W_i^{(x)}x + W_i^{(z)}z_i + b_i\right) (i = 0,1, \dots, L-1). \ (S3)$$

Herein, each element of $W_i^{(x)} \in R^{h \times d}$ is constrained to be non-negative $\left(W_i^{(x)} \geq 0\right)$, while $W_i^{(z)} \in R^{h \times h}$ is unconstrainted, The bias vector is denoted as $b_i \in R^h$, and the activation function is chosen as $\phi(s) = \log(1 + e^x)$. The output is given by:

$$f(x) = \omega^T z_L, \omega \in R^h. (S4)$$

This architecture ensures that $f(x)$ is a convex in $x$ (8).

## Section 3 Benchmark model of two-dimensional energy landscape

We established the benchmark model of two-dimensional energy landscape $V(x_1, x_2)$ below:

$$V(x_1,x_2) = \sum_{i=1}^{3} A_i \exp(-a_i(x_1 - x_{10i})^2 - b_i(x_2 - x_{20i})^2) \quad (S5)$$

Where the parameters are presented as:

$A = [-3 \quad -3 \quad -3];$

$a = [3 \quad 3 \quad 3];$

$b = [3 \quad 3 \quad 3];$

$x_{10} = [-1 \quad 1 \quad 0];$

$x_{20} = [0 \quad 0 \quad 1.5].$

Herein, both Boltzmann constant $k_B$ and temperature T are set to 1.0.

## Section 4 Supplementary Figures

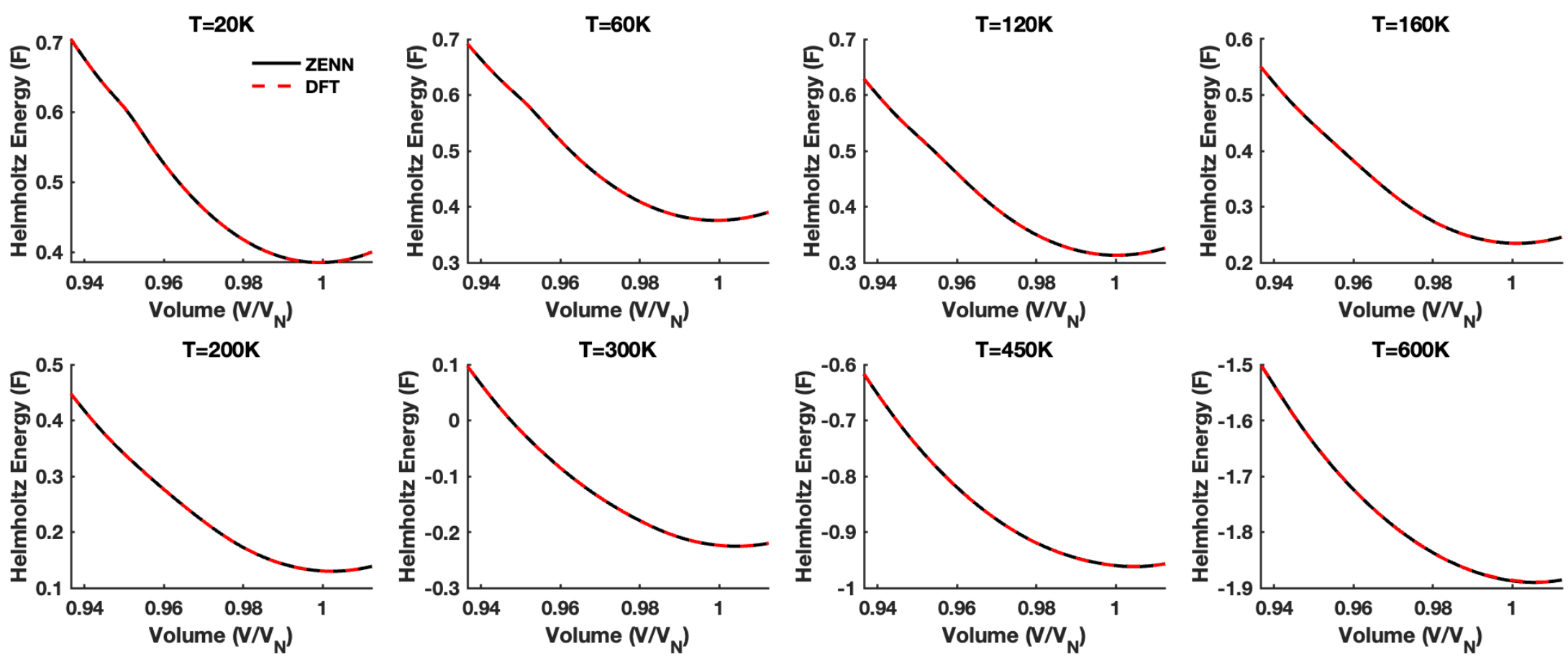

**Fig. S1. Helmholtz energy of $Fe_3Pt$ with the volume $V$ at the different temperatures $T$, with atmospheric pressure.**

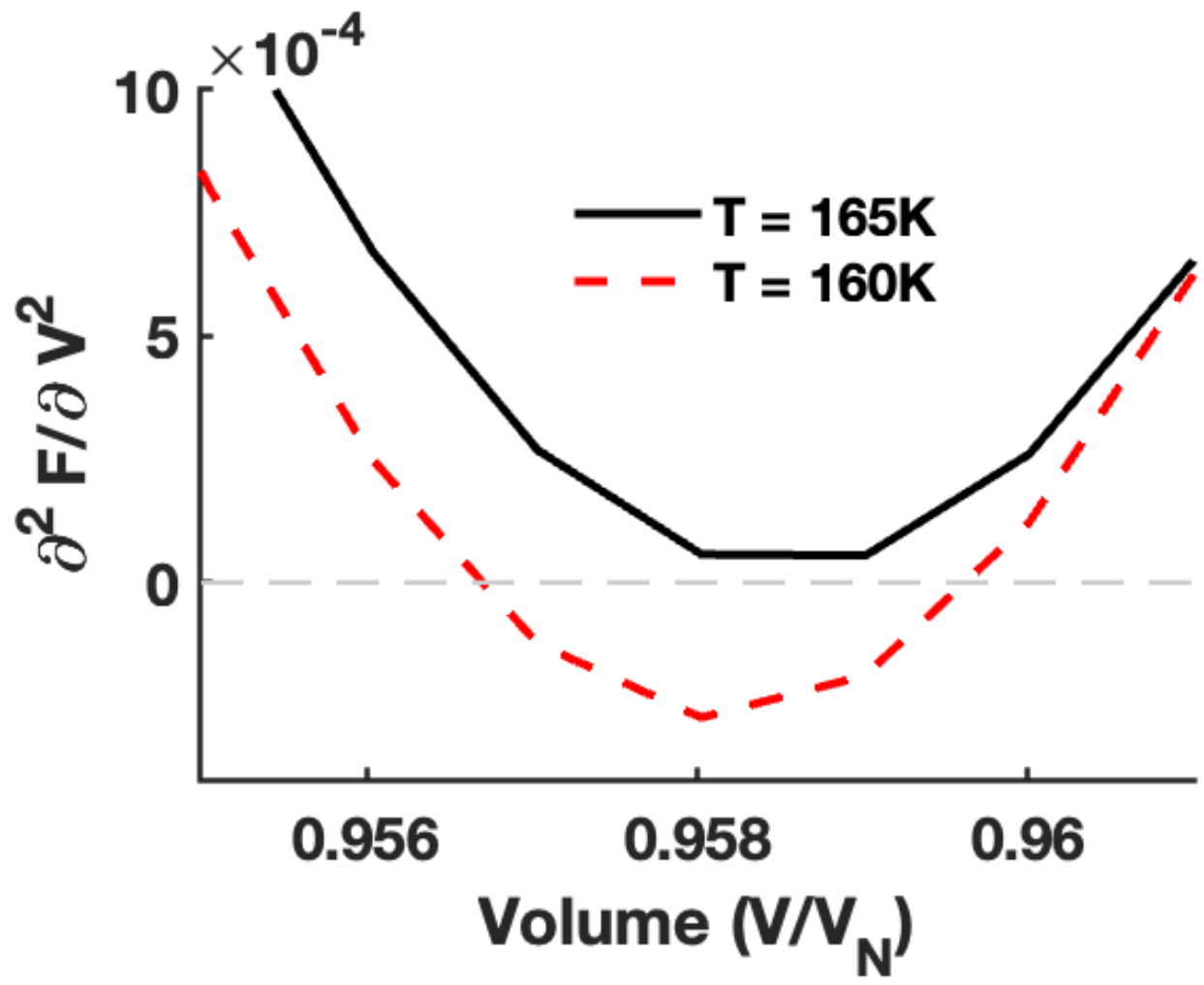

**Fig. S2. Second-order derivative to Helmholtz energy of $\mathrm{Fe_3Pt}$ with the volume $V$ at $T = 160\,K$ and $T = 165\,K$**

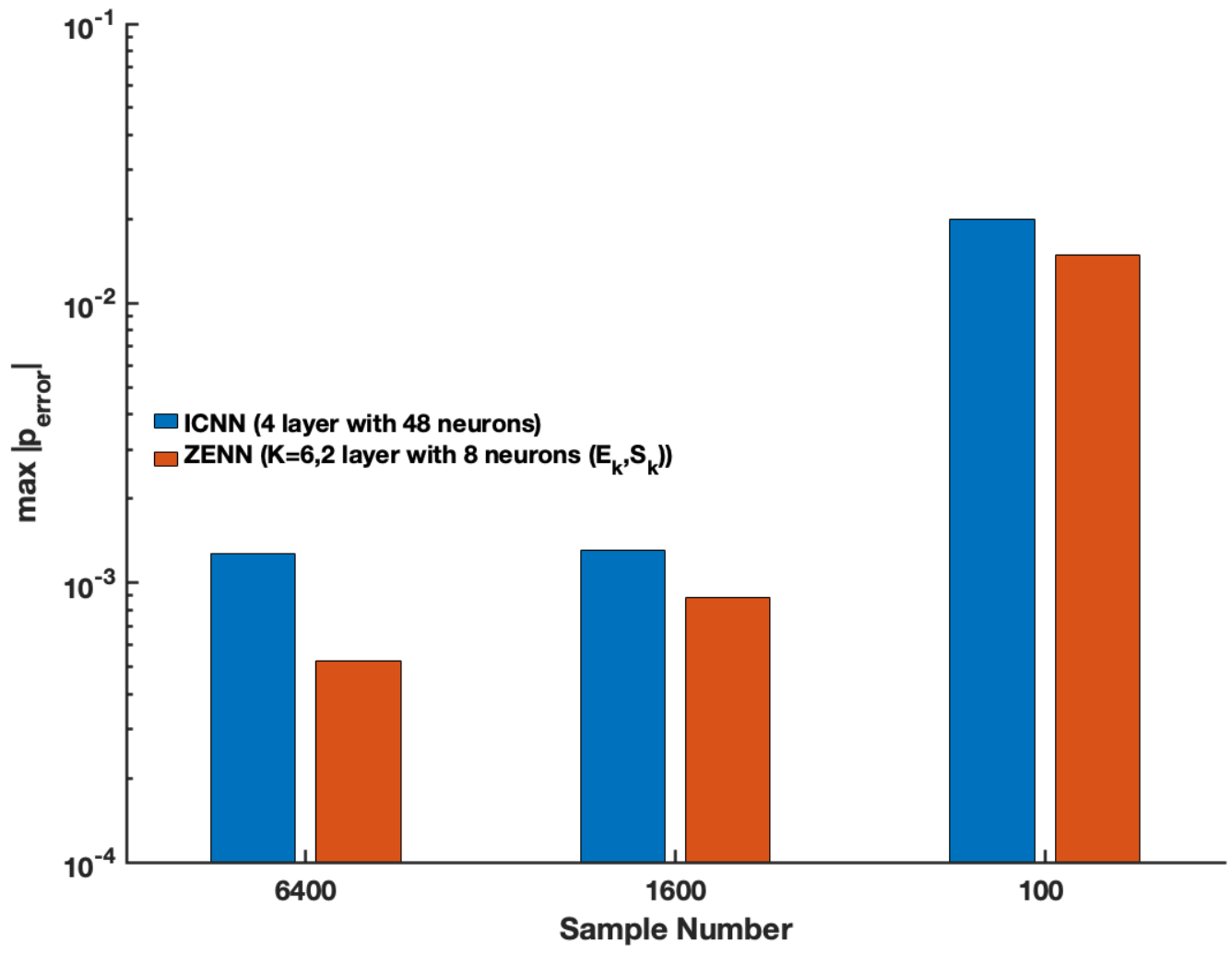

**Fig. S3. Comparison of Generalization Between ICNN and ZENN. $\max|p_{error}|$ denotes the maximum prediction error over the prediction set, which consists of 10,000 samples arranged in a 100×100 grid.**

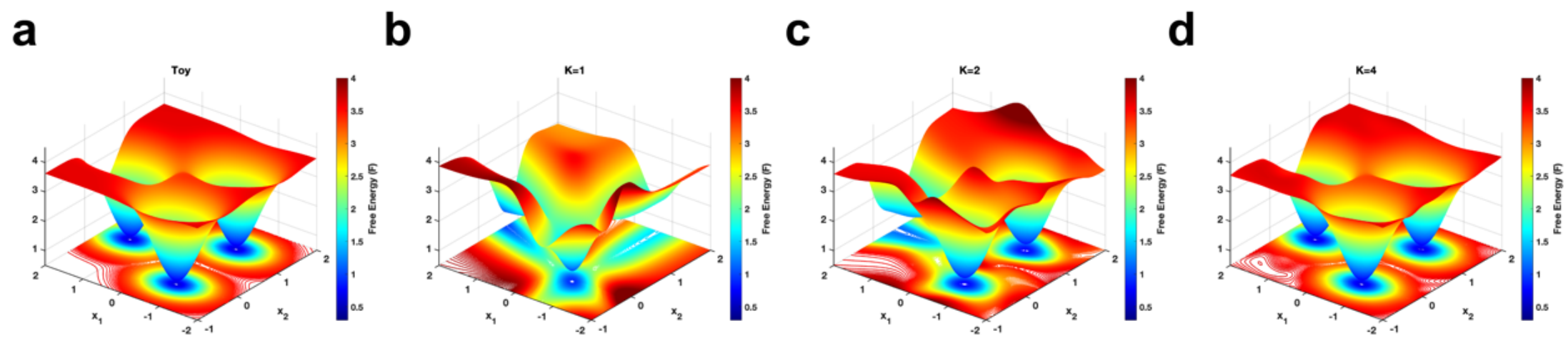

**Fig. S4. Approximation capacity of ZENN with respect to the number of configurations. Benchmark model of two-dimensional energy landscape is presented in section 3.**

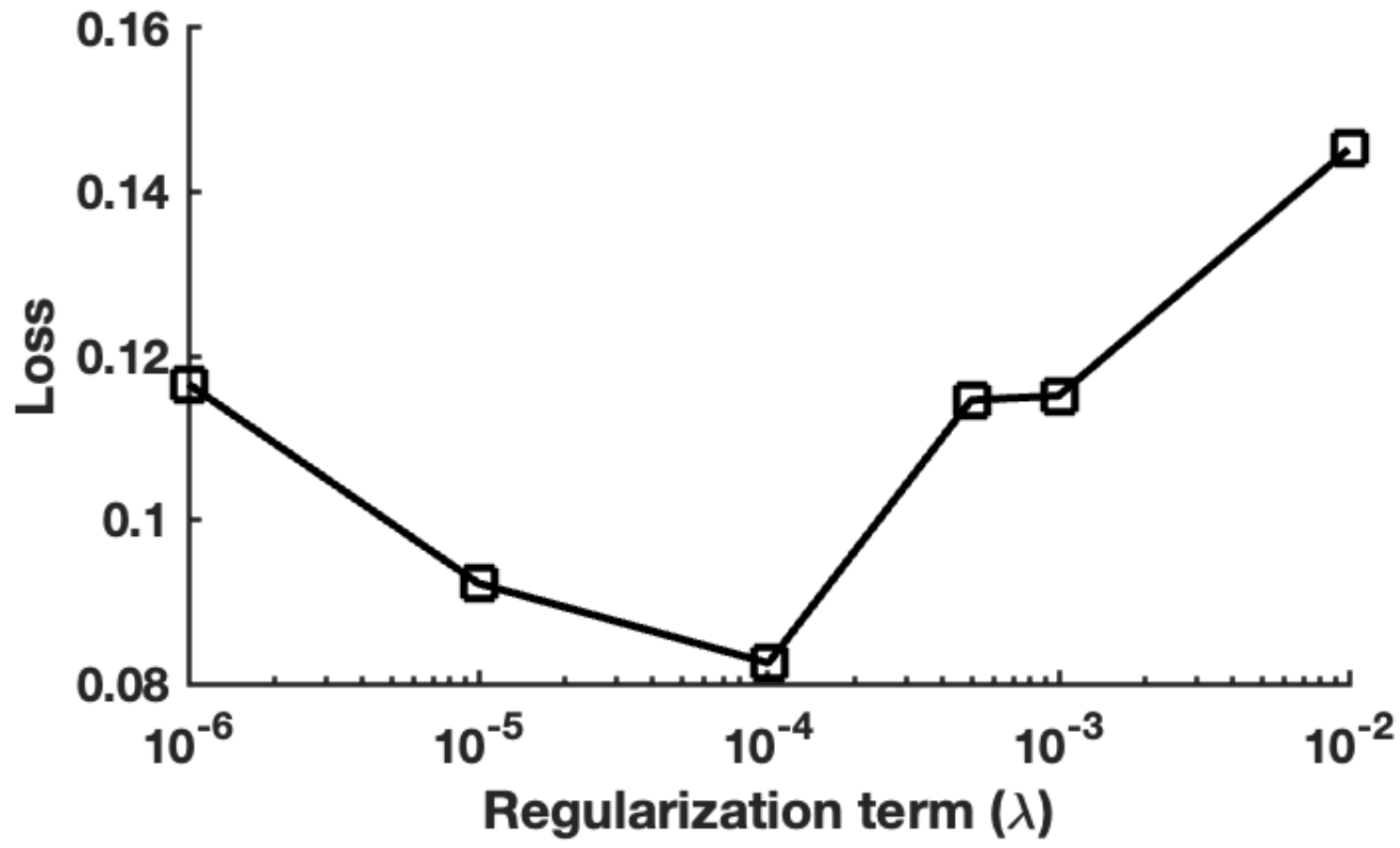

**Fig. S5. Regularization term $\lambda$ associated with the loss function Eq. (20).**

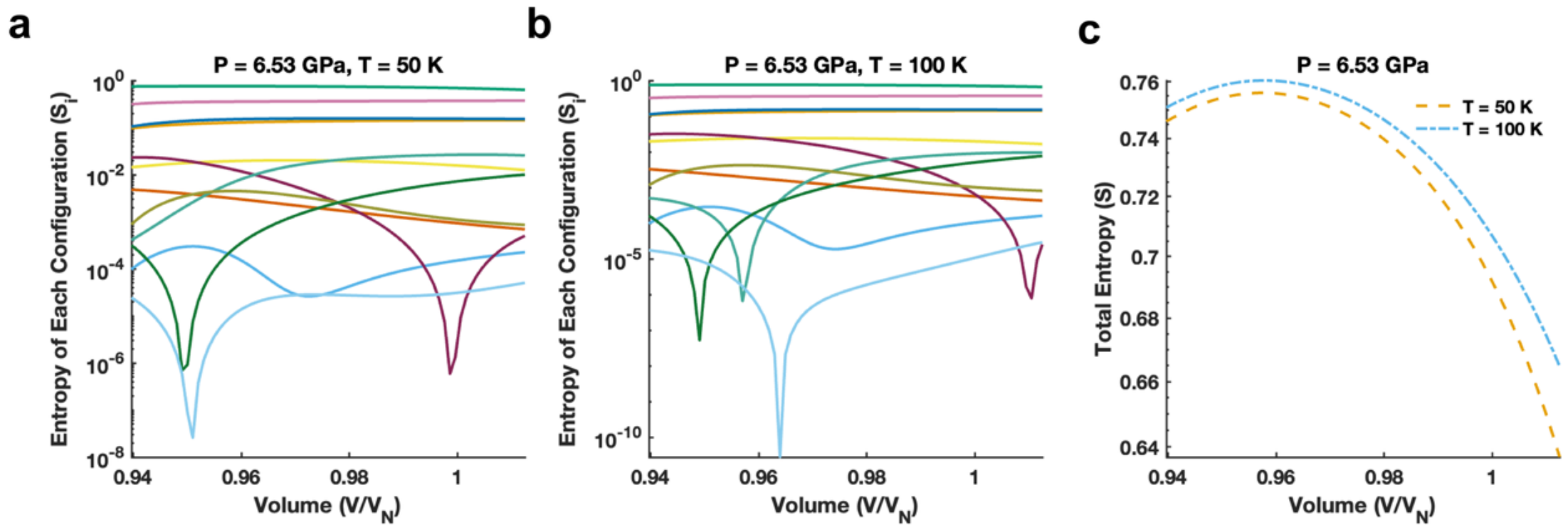

**Fig. S6. Entropy of Fe3Pt computed by ZENN. a. Entropy of each configuration at P = 6.53 GPa and T = 50 K. b. Entropy of each configuration at P = 6.53 GPa and T = 100 K. c. Total Entropy at P = 6.53 GPa. Blue doted and orange dashed line represents T = 100 K and T = 50 K, respectively.**

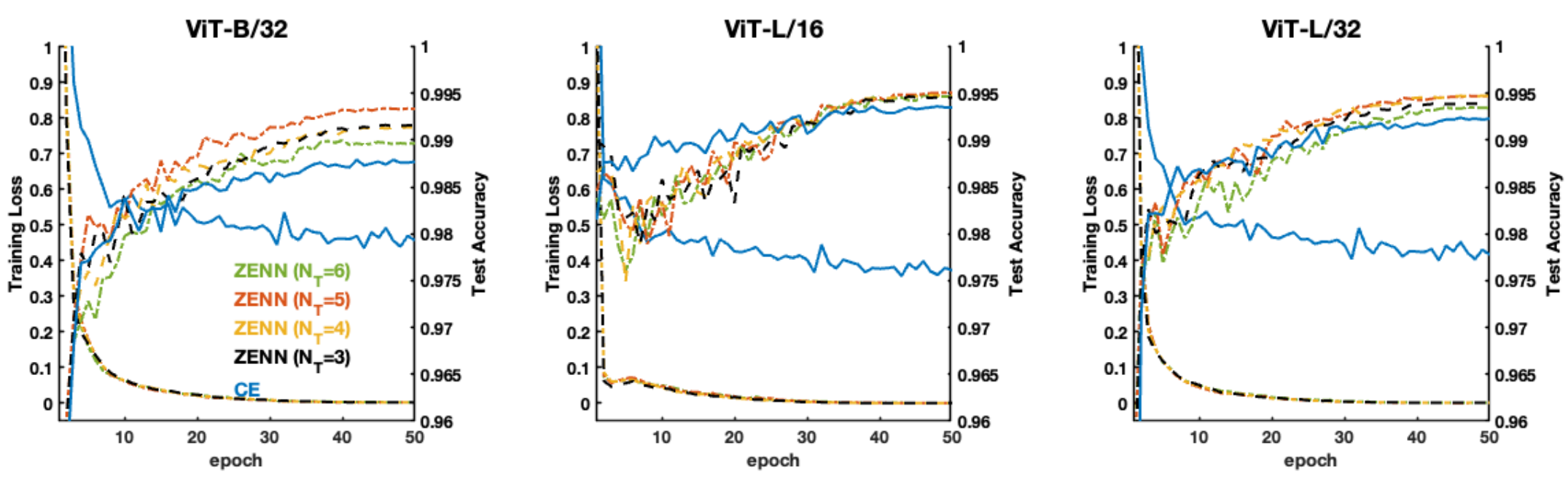

**Fig. S7. Training loss and test accuracy curves for ZENN with different temperature sets compared to CE on the CIFAR-10 dataset.**

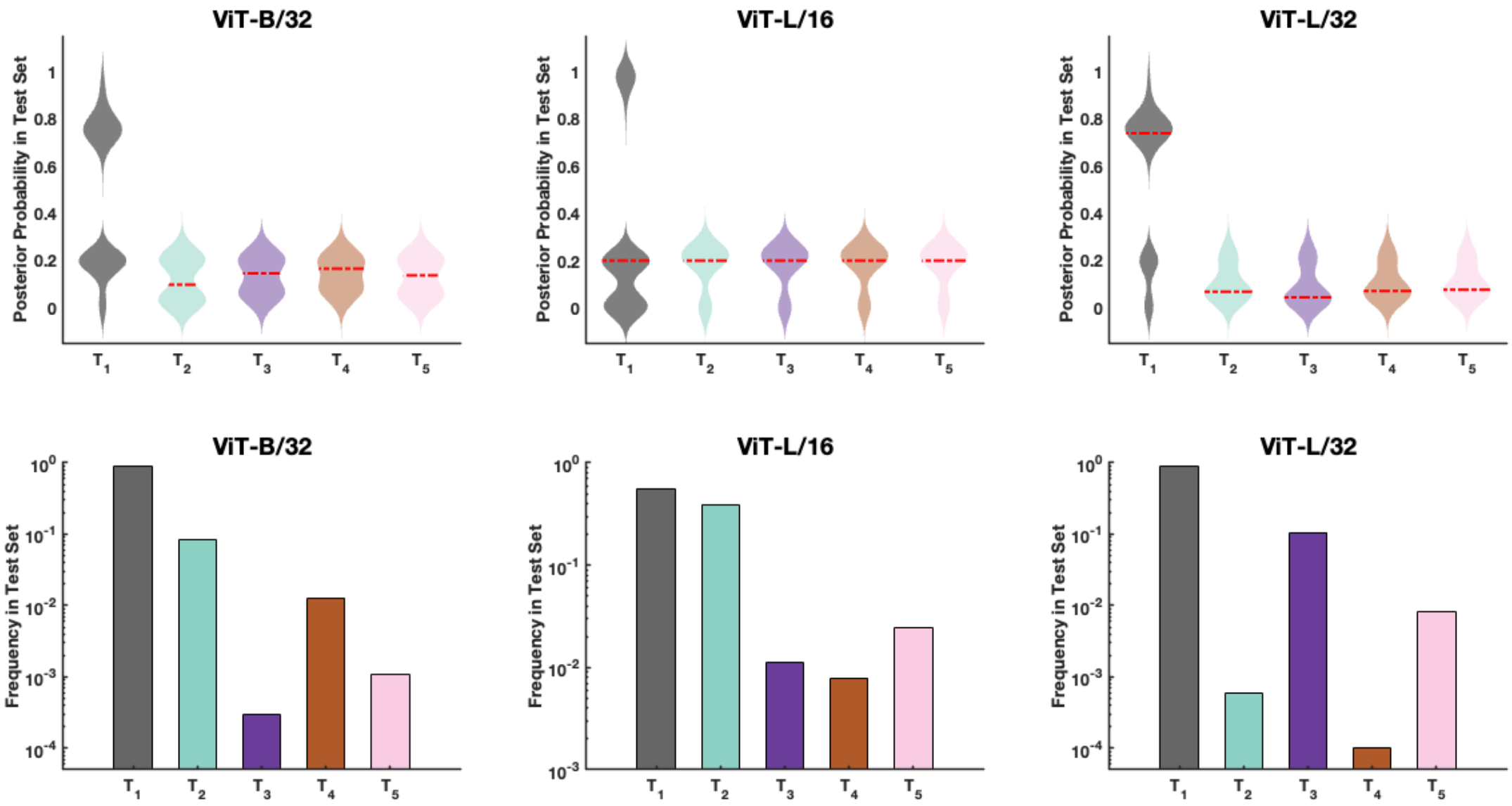

**Fig. S8. Violin plot of the posterior temperature distributions inferred on the test set for $N_T = 5$ on the CIFAR-10 test dataset. The red dashed line indicates the median. The temperature values for various models are listed in Table S12. Frequency distribution of temperature assignments in the CIFAR-10 test set based on one-hot encoding of the MAP temperature.**

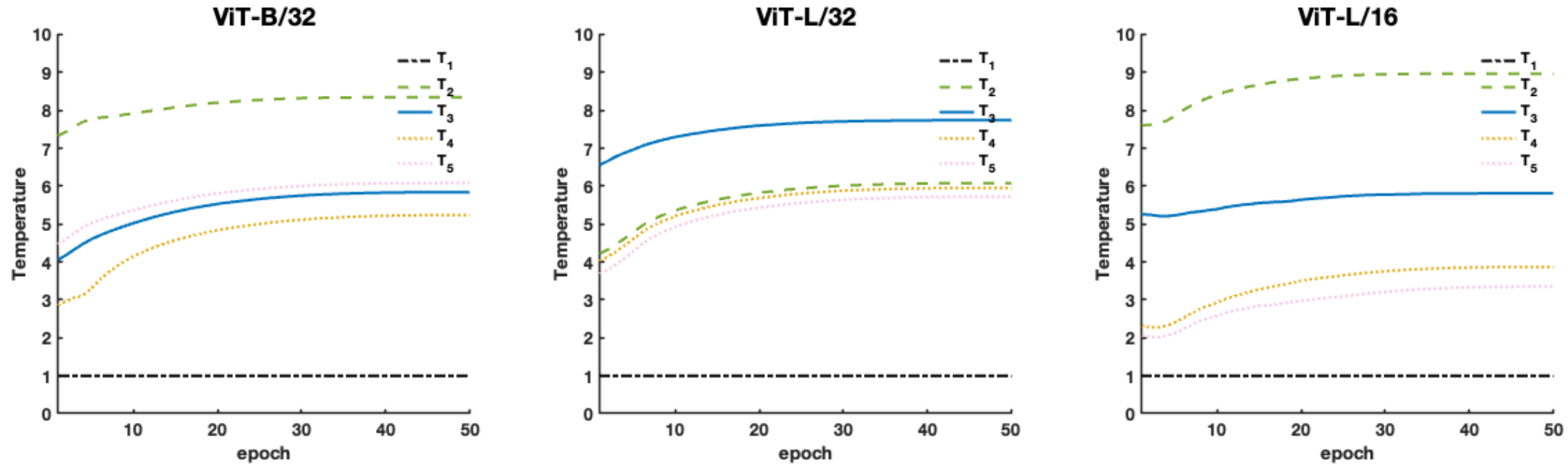

**Fig. S9. Learnable temperature curves for ZENN on the CIFAR-10.**

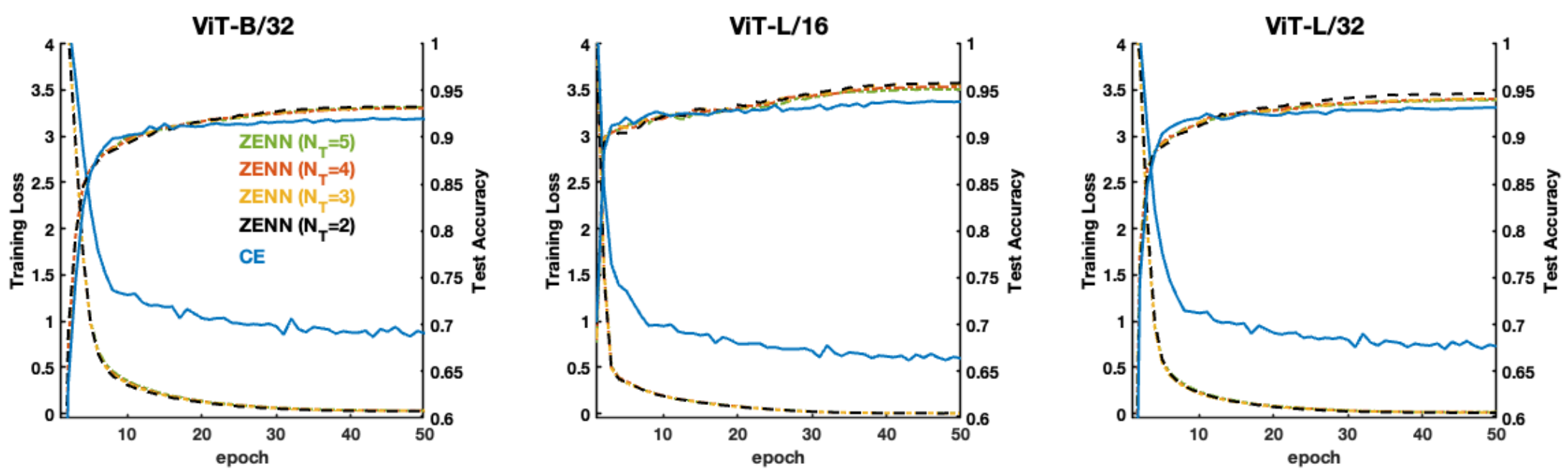

**Fig. S10. Training loss and test accuracy curves for ZENN with different temperature sets compared to CE on the CIFAR-100 dataset.**

.

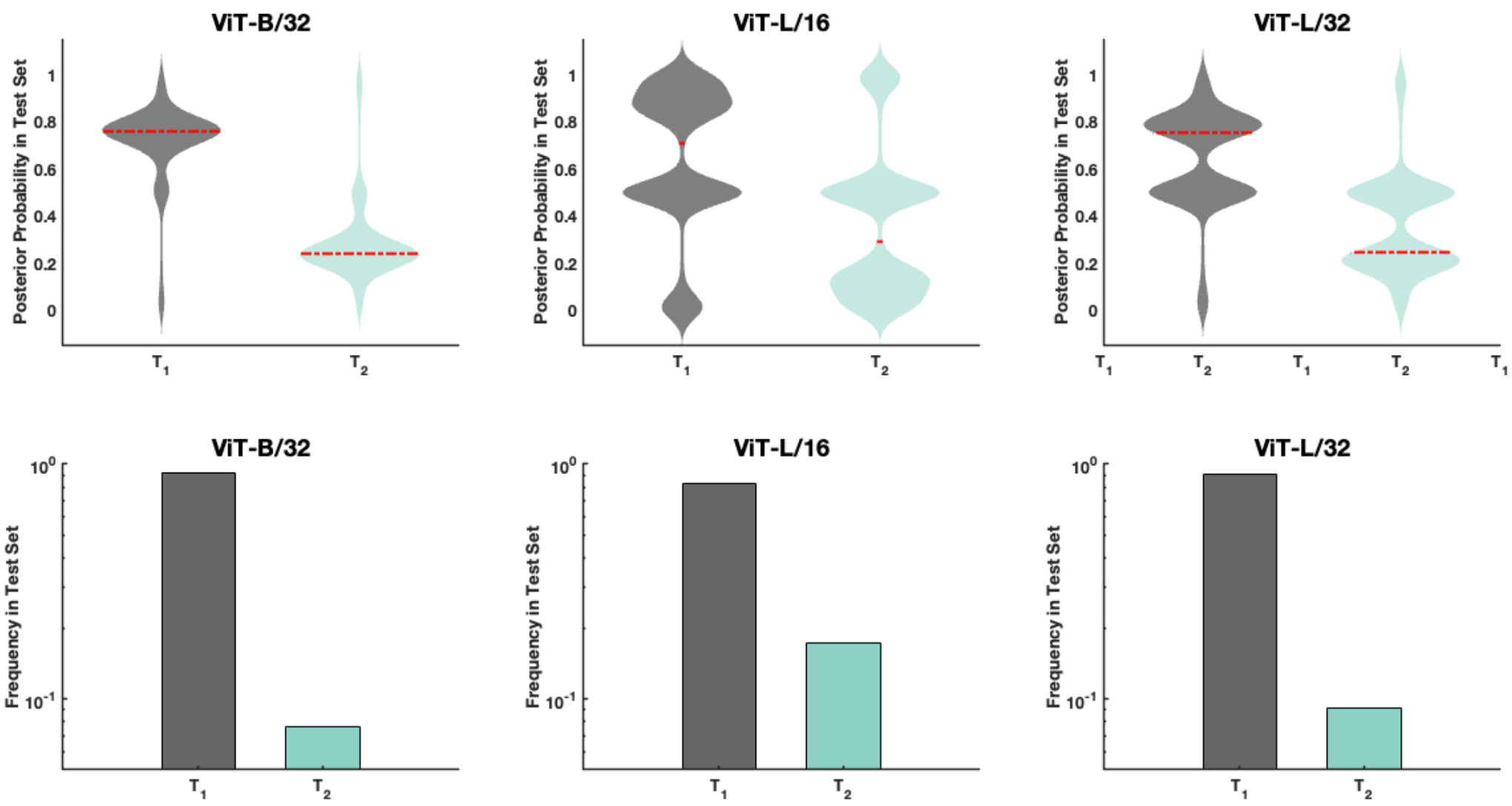

**Fig. S11. Violin plot of the posterior temperature distributions inferred on the test set for $N_T = 2$ on the CIFAR-100 test dataset. The red dashed line indicates the median. The temperature values for various models are listed in Table S13. Frequency distribution of temperature assignments in the CIFAR-100 test set based on one-hot encoding of the MAP temperature.**

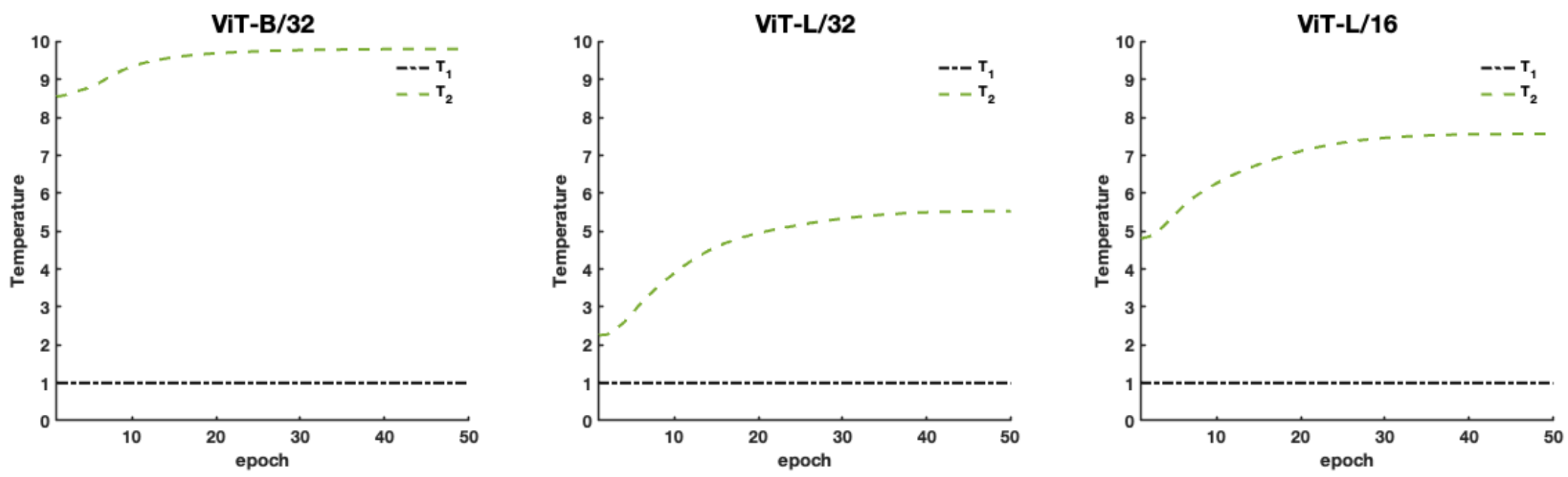

**Fig. S12. Learnable temperature curves for ZENN on the CIFAR-100 datasets.**

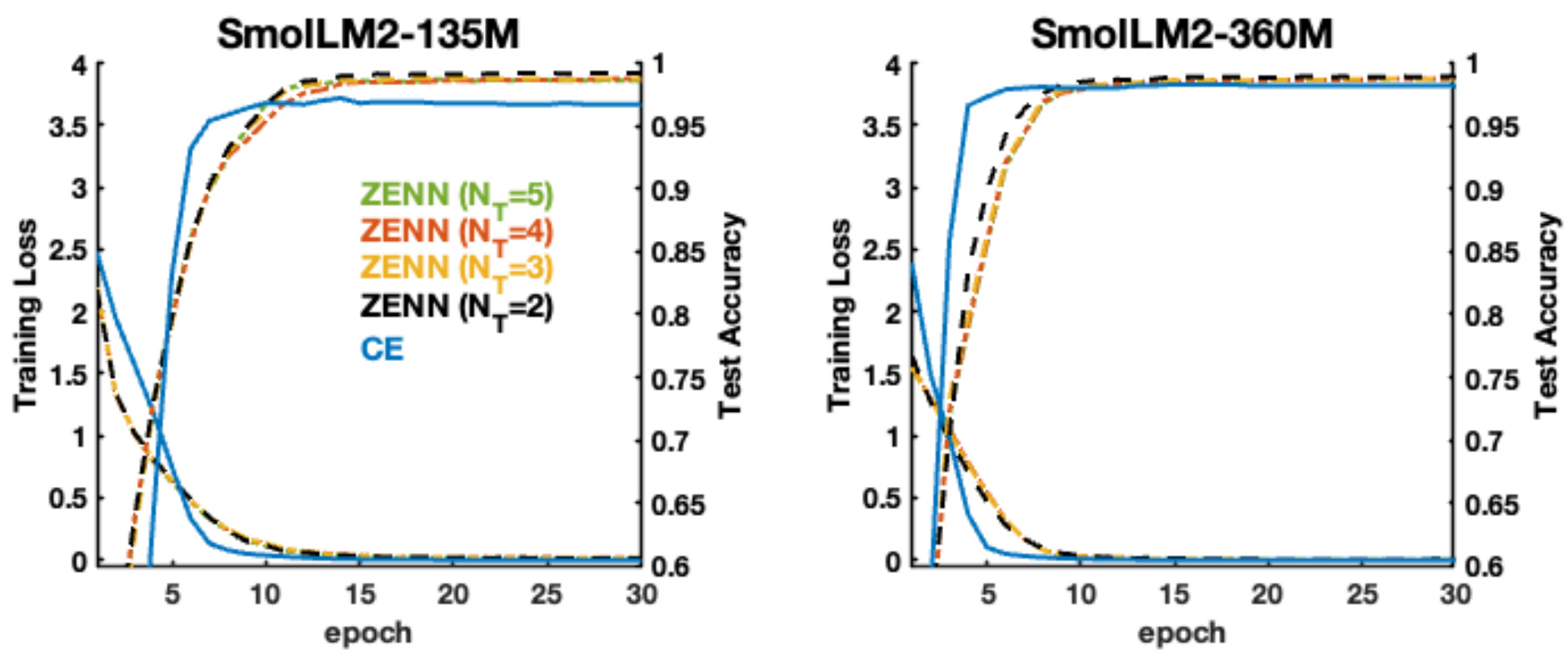

**Fig. S13. Training loss and test accuracy curves for ZENN with different temperature sets compared to CE on the BBCNews dataset.**

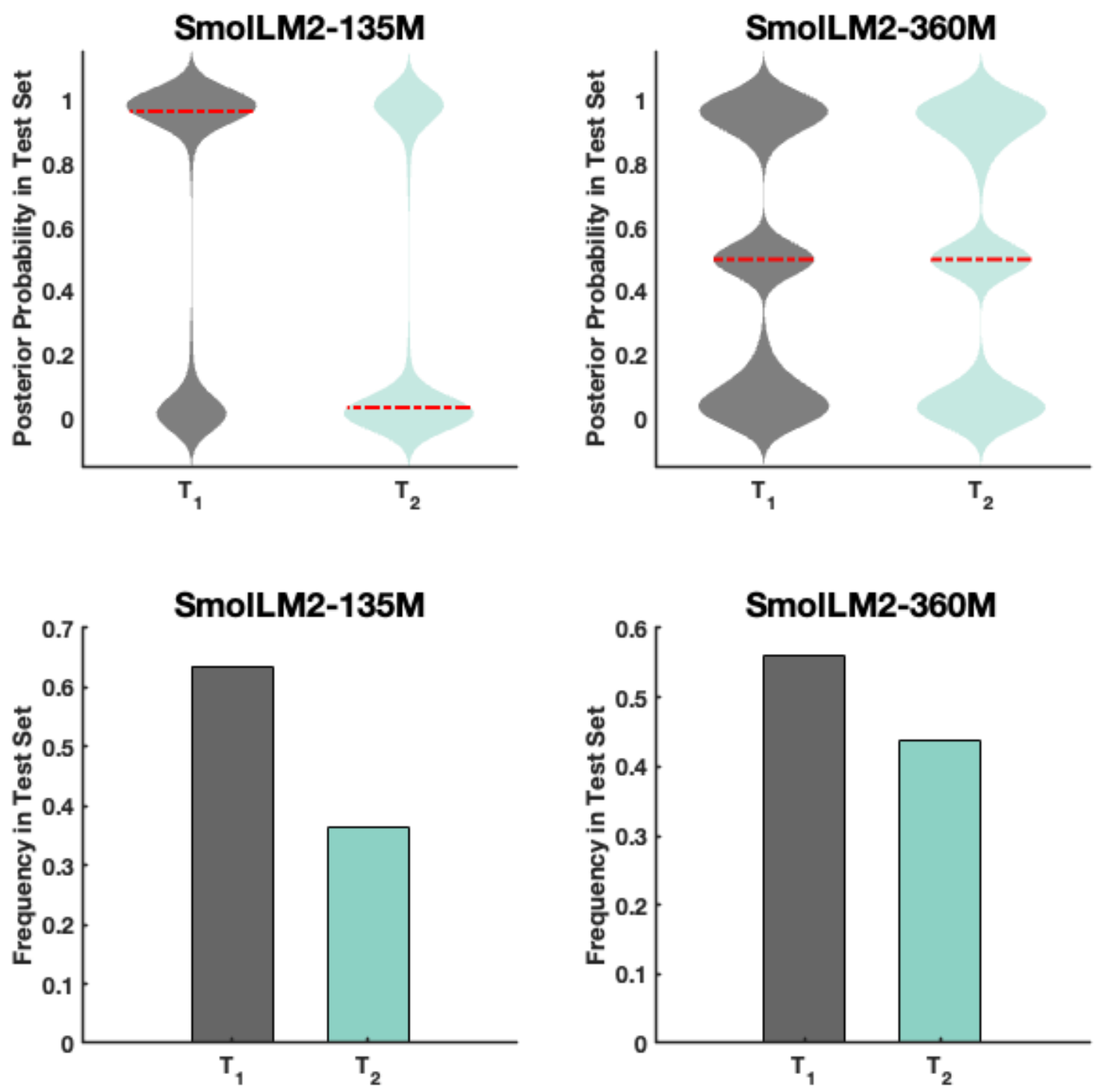

**Fig. S14. Violin plot of the posterior temperature distributions inferred on the test set for $N_T = 2$ on the BBCNews test dataset. The red dashed line indicates the median. The temperature values for various models are listed in Table S14. Frequency distribution of temperature assignments in the BBCNews test set based on one-hot encoding of the MAP temperature.**

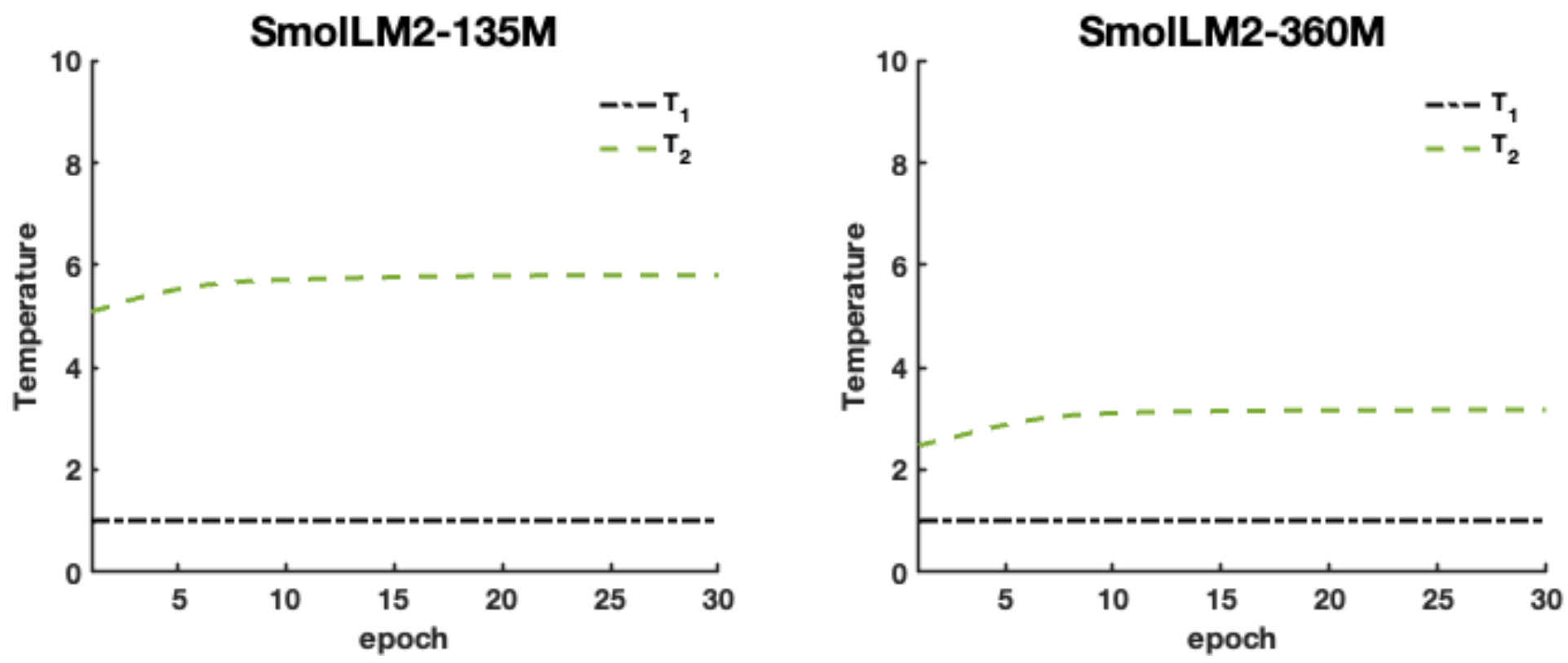

**Fig. S15. Learnable temperature curves for ZENN on the BBCNews datasets.**

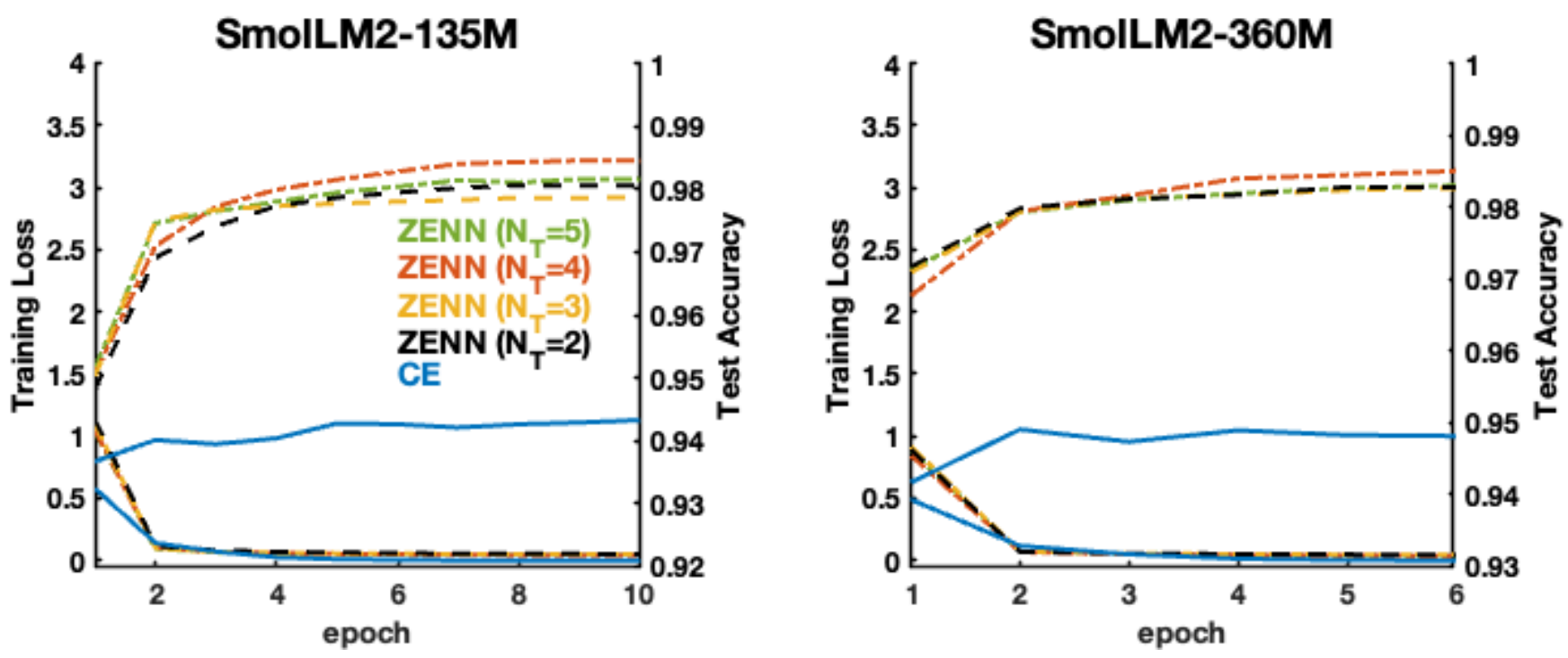

**Fig. S16. Training loss and test accuracy curves for ZENN with different temperature sets compared to CE on the AGNews dataset.**

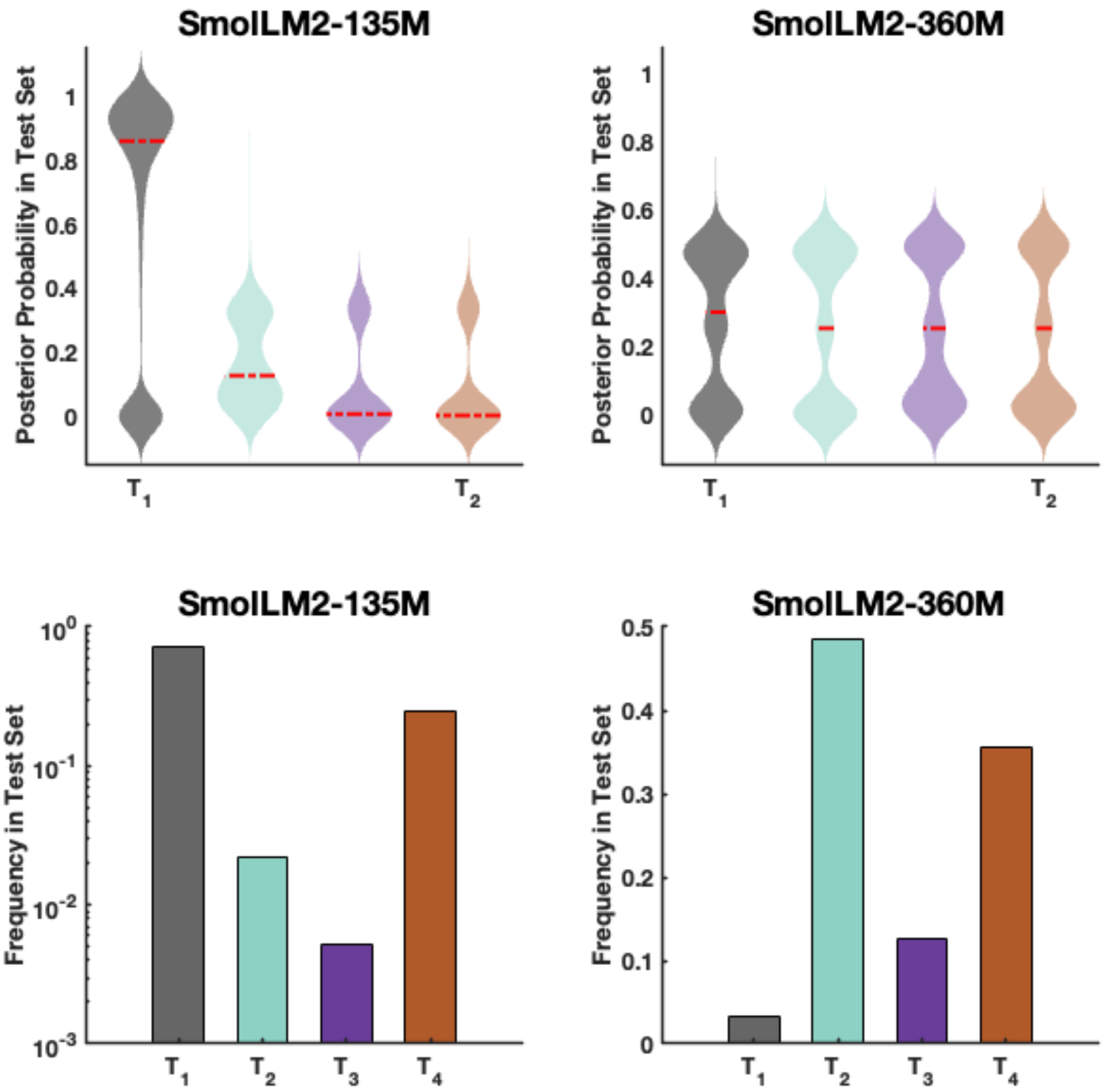

**Fig. S17. Violin plot of the posterior temperature distributions inferred on the test set for $N_T = 4$ on the AGNews test dataset. The red dashed line indicates the median. The temperature values for various models are listed in Table S15. Frequency distribution of temperature assignments in the AGNews test set based on one-hot encoding of the MAP temperature.**

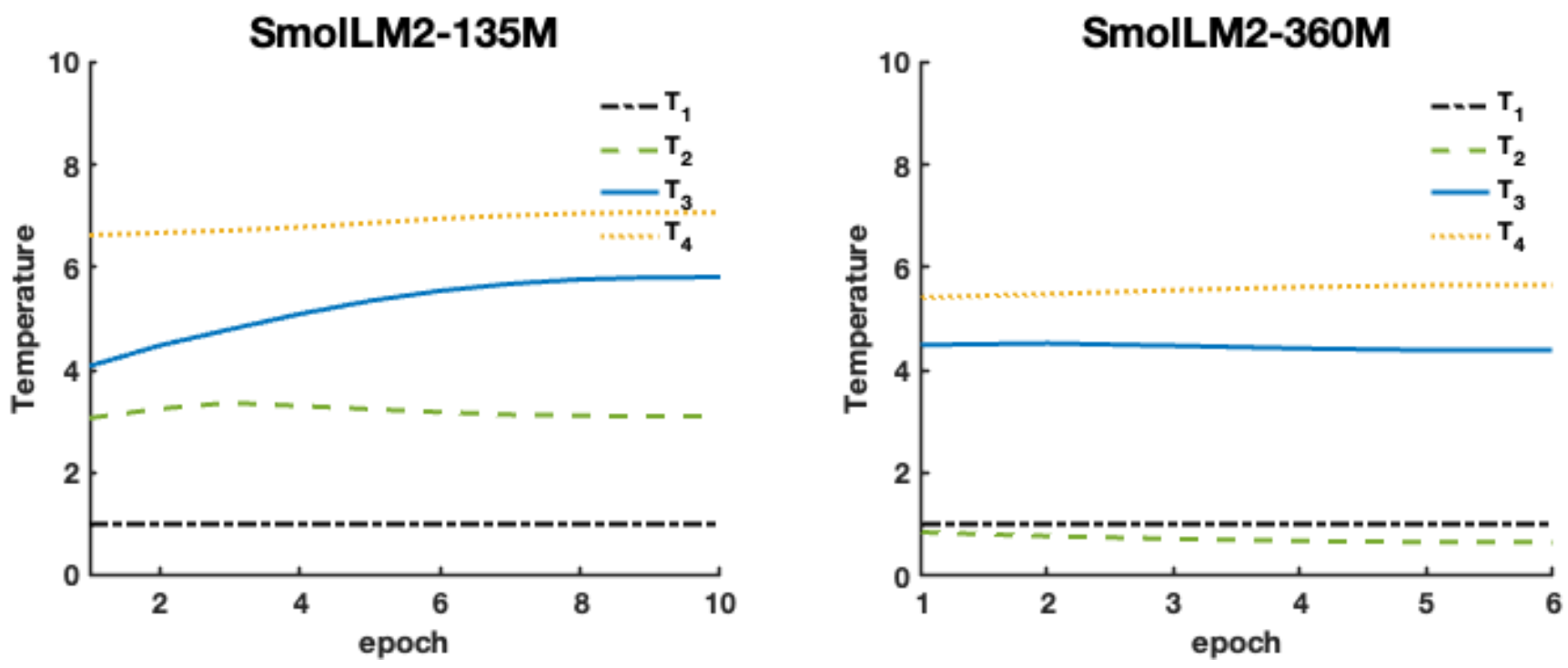

**Fig. S18. Learnable temperature curves for ZENN on the AGNews datasets.**

## Section 5 Supplementary Tables

**Table S1**. Fitted equilibrium properties by Eq. (S2) for $Fe_3Pt$ using the 12-atom supercell, including $V_0$, $E_0$, $B_0$, and $B'$. Here, configuration #1 is the ferromagnetic configuration, "DF" indicates degeneracy factor for each configuration with the total degeneracy factors of 512 ($2^9$) due to 9 spin up and spin down Fe atoms. Note that these results are also reported in (1).

| Configuration | DF | $V_0$ (Å$^3$/atom) | $E_0$ (eV/atom) | $B_0$ (GPa) | $B'$ |
|---|---|---|---|---|---|
| 1 | 2 | 13.124 | 0.0000 | 177.01 | 3.447 |
| 2 | 6 | 12.901 | 0.0122 | 163.06 | 4.260 |
| 3 | 12 | 12.910 | 0.0160 | 162.96 | 4.440 |
| 4 | 12 | 13.002 | 0.0191 | 171.16 | 3.501 |
| 5 | 6 | 13.005 | 0.0202 | 171.75 | 3.271 |
| 6 | 6 | 12.929 | 0.0211 | 164.05 | 3.740 |
| 7 | 24 | 12.890 | 0.0281 | 162.43 | 3.923 |
| 8 | 24 | 12.947 | 0.0291 | 164.33 | 3.159 |
| 9 | 12 | 12.875 | 0.0372 | 164.70 | 4.059 |
| 10 | 12 | 12.868 | 0.0419 | 170.28 | 4.144 |
| 11 | 12 | 12.867 | 0.0422 | 171.76 | 3.721 |
| 12 | 24 | 12.840 | 0.0455 | 161.43 | 3.772 |
| 13 | 6 | 12.767 | 0.0457 | 161.03 | 4.388 |
| 14 | 12 | 12.859 | 0.0462 | 163.01 | 3.224 |
| 15 | 12 | 12.864 | 0.0467 | 168.31 | 4.357 |
| 16 | 24 | 12.801 | 0.0491 | 167.54 | 4.154 |
| 17 | 24 | 12.804 | 0.0493 | 165.12 | 4.649 |
| 18 | 12 | 12.765 | 0.0494 | 157.04 | 4.507 |
| 19 | 24 | 12.789 | 0.0505 | 169.38 | 4.816 |
| 20 | 12 | 12.833 | 0.0517 | 169.08 | 4.029 |
| 21 | 6 | 12.834 | 0.0521 | 164.59 | 3.891 |
| 22 | 6 | 12.747 | 0.0533 | 163.93 | 4.670 |
| 23 | 24 | 12.765 | 0.0551 | 159.41 | 4.137 |
| 24 | 12 | 12.724 | 0.0559 | 170.18 | 4.298 |
| 25 | 24 | 12.715 | 0.0586 | 172.04 | 4.622 |
| 26 | 24 | 12.779 | 0.0593 | 167.06 | 4.309 |
| 27 | 24 | 12.788 | 0.0596 | 170.54 | 4.187 |
| 28 | 12 | 12.727 | 0.0599 | 164.79 | 4.109 |
| 29 | 24 | 12.806 | 0.0623 | 169.21 | 4.052 |
| 30 | 12 | 12.703 | 0.0686 | 174.20 | 4.283 |
| 31 | 12 | 12.729 | 0.0712 | 169.24 | 4.215 |
| 32 | 12 | 12.740 | 0.0733 | 165.45 | 4.122 |
| 33 | 12 | 12.716 | 0.0877 | 163.09 | 3.917 |
| 34 | 12 | 12.754 | 0.0895 | 162.97 | 3.933 |
| 35 | 2 | 12.684 | 0.0899 | 167.31 | 4.483 |
| 36 | 4 | 12.681 | 0.0900 | 167.39 | 4.735 |
| 37 | 12 | 12.721 | 0.0907 | 162.57 | 4.081 |

**Table S2**. Hyperparameters in ZENN.

| | $k_B$ | $\gamma$ |
|---|---|---|
| Classification | 1.0 | 100.0 |
| Energy landscape | 1.0 | 5.0 |
| $Fe_3Pt$ | 0.1 | 5.0 |

**Table S3** Hyperparameters for Vision Transformer (ViT) in ZENN on CIFAR-10 & CIFAR-100

| | |
|---|---|
| Data Augmentation | {RandomHorizontalFlip & RandomCrop} |
| Number of epochs | 50 |
| Resolution | 224 |
| Batch size | 64 |
| Initial learning rate | $1 \times 10^{-4}$ |
| Initial learning rate of temperature | $1 \times 10^{-3}$ |
| Weight decay | 0.1 |
| Optimizer | AdamW |
| Pretrained Weights | ImageNet-21k pretrained weights |

Note: More details are presented on shell script on Github.

**Table S4** Hyperparameters for Vision Transformer (ViT) in CE on CIFAR-10 & CIFAR-100

| | |
|---|---|
| Data Augmentation | {RandomHorizontalFlip & RandomCrop} |
| Number of epochs | 50 |
| Resolution | 224 |
| Batch size | 64 |
| Initial learning rate | $1 \times 10^{-4}$ |
| Weight decay | 0.1 |
| Optimizer | AdamW |
| Pretrained Weights | ImageNet-21k pretrained weights |

Note: More details are presented on shell script on Github.

**Table S5** Hyperparameters for SmolLM2 in ZENN on BBC News

| | |
|---|---|
| Number of epochs | 30 |
| Maximum Length | 512 |
| Batch size | 32 |
| Initial learning rate | $5 \times 10^{-5}$ |
| Initial learning rate of temperature | $1 \times 10^{-3}$ |
| Weight decay | 0.01 |
| Optimizer | AdamW |
| Fine-Tune | LoRA $(r = 8, \alpha = 16, \text{dropout} = 0.1)$ q_proj, v_proj |

Note: More details are presented on shell script on Github.

**Table S6** Hyperparameters for SmolLM2 in CE on BBC News

| | |
|---|---|
| Number of epochs | 30 |
| Maximum Length | 512 |
| Batch size | 32 |
| Initial learning rate | $5 \times 10^{-5}$ |
| Weight decay | 0.01 |
| Optimizer | AdamW |
| Fine-Tune | LoRA $(r = 8, \alpha = 16, \text{dropout} = 0.1)$ q_proj, v_proj |

Note: More details are presented on shell script on Github.

**Table S7** Hyperparameters for SmolLM2 in ZENN on AG News

| | |
|---|---|
| Number of epochs | 6 |
| Maximum Length | 512 |
| Batch size | 32 |
| Initial learning rate | $5 \times 10^{-5}$ |
| Initial learning rate of temperature | $1 \times 10^{-3}$ |
| Weight decay | 0.01 |
| Optimizer | AdamW |
| Fine-Tune | LoRA $(r = 8, \alpha = 16, \text{dropout} = 0.1)$ q_proj, v_proj |

Note: More details are presented on shell script on Github.

**Table S8** Hyperparameters for SmolLM2 in CE on AG News

| | |
|---|---|
| Number of epochs | 10 |
| Maximum Length | 512 |
| Batch size | 32 |
| Initial learning rate | $5 \times 10^{-5}$ |
| Weight decay | 0.01 |
| Optimizer | AdamW |
| Fine-Tune | all parameters |

Note: More details are presented on shell script on Github.

**Table S9** Trainable Parameter Scale for ZENN and CE on CIFAR datasets

| | ZENN | CE |
|---|---|---|
| ViT-B/32 | 174.9M | 87.5M |
| ViT-L/16 | 606.6M | 303.3M |
| ViT-L/32 | 611M | 305.5M |

**Table S10** Trainable Parameter Scale for ZENN and CE on BBC News

| | ZENN | CE |
|---|---|---|
| SmolLM2-135M | 0.9M | 0.5M |
| SmolLM2-360M | 1.6M | 0.8M |

**Table S11** Trainable Parameter Scale for ZENN and CE on AG News

| | ZENN | CE |
|---|---|---|
| SmolLM2-135M | 0.9M | 135M |
| SmolLM2-360M | 1.6M | 360M |

**Table S12** Trainable Temperature for ZENN on CIFAR-10

| | $T_1$ | $T_2$ | $T_3$ | $T_4$ | $T_5$ |
|---|---|---|---|---|---|
| ViT-B/32 | 1.0 | 8.345 | 5.842 | 5.237 | 6.083 |
| ViT-L/32 | 1.0 | 6.081 | 7.741 | 5.954 | 5.723 |
| ViT-L/16 | 1.0 | 8.969 | 5.809 | 3.869 | 3.353 |

**Table S13** Trainable Temperature for ZENN on CIFAR-100

| | $T_1$ | $T_2$ |
|---|---|---|
| ViT-B/32 | 1.0 | 9.808 |
| ViT-L/32 | 1.0 | 5.529 |
| ViT-L/16 | 1.0 | 7.567 |

**Table S14** Trainable Temperature for ZENN on BBC News

| | $T_1$ | $T_2$ |
|---|---|---|
| SmolLM2-135M | 1.0 | 5.807 |
| SmolLM2-360M | 1.0 | 3.163 |

**Table S15** Trainable Temperature for ZENN on AG News

| | $T_1$ | $T_2$ | $T_3$ | $T_4$ |
|---|---|---|---|---|
| SmolLM2-135M | 1.0 | 3.095 | 5.811 | 7.075 |
| SmolLM2-360M | 1.0 | 0.643 | 4.383 | 5.656 |